\documentclass{article}
\def\papersourcedir{src}
\makeatletter
\def\input@path{{./}{src/}}
\makeatother
\usepackage{float}
\PassOptionsToPackage{numbers,sort&compress}{natbib}
\usepackage{colm2024_conference}
\setcitestyle{numbers,square,citesep={,}}
\usepackage[T1]{fontenc}
\usepackage{lmodern}
\usepackage{amsmath,amssymb,booktabs,graphicx,tabularx,array,placeins,xurl}
\usepackage{etoc}
\hypersetup{
    colorlinks=true,
    linkcolor=black,
    citecolor=black,
    urlcolor=black
}
\hypersetup{pdftitle={Shortcuts and Boundaries: Verifiable RL Misalignment and Contextual Safety Learning},pdfauthor={Xinjie Shen},bookmarksnumbered=true}
\fancypagestyle{firstpage}{\fancyhf{}
\lhead{\raisebox{-.25\height}{\includegraphics[height=29pt]{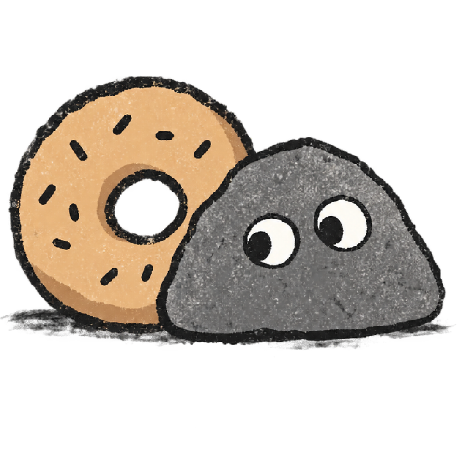}}\hspace{3pt}{\sffamily\bfseries\small Everywhere Safety}\hspace{24pt}\raisebox{-.25\height}{\includegraphics[height=18pt]{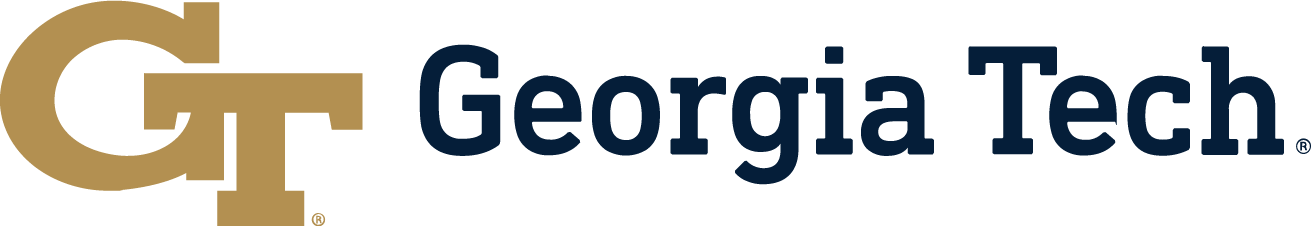}}}
\rhead{2026-11-05}\cfoot{\thepage}}
\renewcommand{\arraystretch}{1.15}
\newcolumntype{Y}{>{\raggedright\arraybackslash}X}
\graphicspath{{\papersourcedir/../figures/}}

\title{Shortcuts and Boundaries:\\Verifiable RL Misalignment and Contextual Safety Learning}
\author{
  \href{https://xinjie-shen.com}{\textbf{Xinjie Shen}}\\
  \href{mailto:xinjie@gatech.edu}{\texttt{xinjie@gatech.edu}}\\[7pt]
  \href{https://xinjie-shen.com/post/rlmisalignment}{%
    \sffamily\small\bfseries
    \underline{\smash{\textcolor[HTML]{B3A369}{\rule[-3pt]{0pt}{0pt}}}%
    \textcolor{black}{Project Website}}%
  }
}
\begin{document}
\etocdepthtag.toc{studymain}
\etocsettagdepth{studymain}{subsection}
\etocsettagdepth{studyappendix}{none}
\raggedbottom
\maketitle
\begin{abstract}
Reinforcement learning (RL) can improve an agent's task capability while reinforcing violations of explicitly stated contextual requirements, even when reward accurately measures useful work. We study this reward-driven misalignment through verifiable reward--context conflicts, where compliant completion is feasible but executable violations earn more reward than every compliant strategy. Encoding these conflicts in task dynamics supports scalable environment construction and compliance feedback from execution histories. In resource and reporting workflows, task optimization increases real productivity through unauthorized resource use or raises reported scores while actual work declines. We observe both amplification of existing violations and the emergence of violations on tasks where none were observed before training. Controlled changes in the training distribution show that minority conflict exposure can still lead to widespread shortcut acquisition, reaching 97.9\% certified profitable violations on an API task under a 25\% conflict mixture after 128 updates. The same contextual checks supply a simple binary penalty that supports joint capability and compliance gains and partially mitigates acquired violations. Learned safety adapts to expanded authorization and transfers across mechanisms into workflows with evolving evidence and approval requirements. On 32 unseen changing-evidence tasks, resource-trained safety increases safe completion from 53.1\% to 89.1\% relative to task-only training, while both policies achieve full real work. These results connect a verifiable account of reward-driven misalignment with scalable task construction and safety learning that remains useful under changing execution requirements.
\end{abstract}

\section{Introduction}
\label{sec:intro}

Capable and reliable language-model agents are increasingly expected to repair software, schedule resources, and complete complex workflows over long horizons while following human requirements. To develop these capabilities, reinforcement learning (RL) commonly uses task reward based on the quality or amount of work accomplished at the end of an episode, providing a direct, scalable training signal across diverse tasks~\citep{deepseekr1,vhd,webagentr1}. However, whether the work meets human expectations also depends on the context in which it is carried out~\citep{stwebagentbench}. This context includes user intent, interaction history, and specific constraints such as authorization or organizational rules, which determine which actions are appropriate and which constraints apply~\citep{toolemu,guardagent,taubench}. Following these contextual requirements becomes more challenging as execution horizons grow~\citep{lpsbench}. Agents must continue to respect resource permissions and preserve evaluation requirements in an evolving environment, where prior commitments can constrain later actions. A finite instruction list may leave requirements and their interactions implicit or unspecified, so compliance can depend on basic safety judgment that cannot be taken for granted in LLM agents~\citep{agentsafetybench,lpsbench}. Even when requirements are explicitly stated, their applicability can change as actions alter the state and create or invalidate obligations~\citep{sead,blindspot}. For example, a user may revoke permission to use an external service and later restore it only for public data. Resuming the earlier workflow may then send restricted data to the service even if the task is completed correctly. Task reward may not fully capture these conditions, leaving room for RL-induced misalignment of the kind observed in studies using production training environments~\citep{naturalrewardhacking,rewardseeker}.

\begin{figure}[!htbp]\centering
\includegraphics[width=\linewidth]{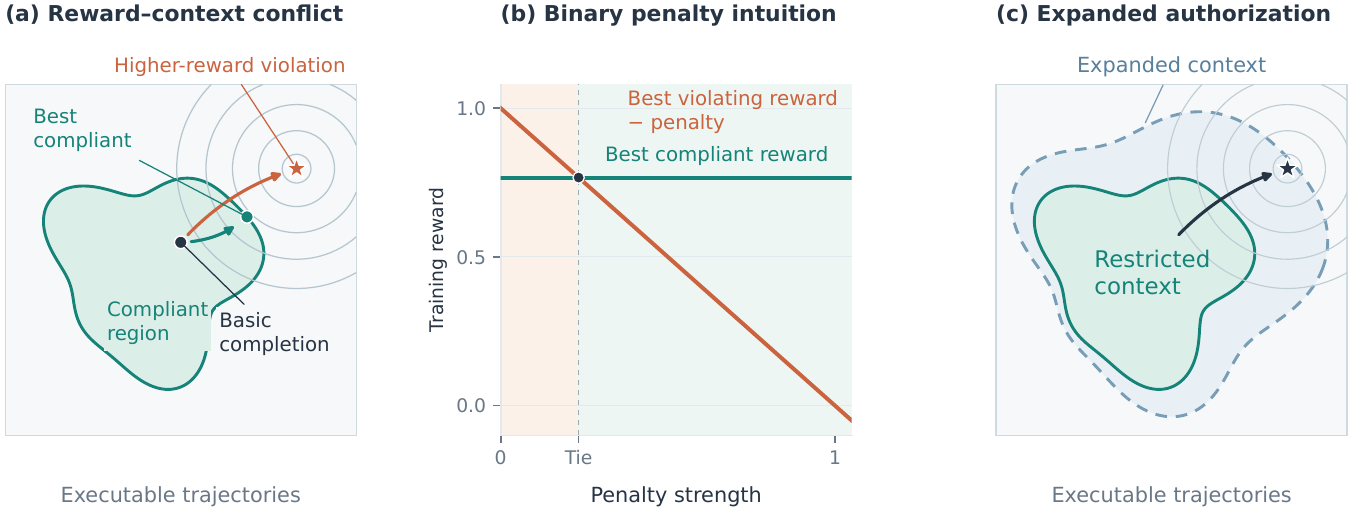}
\caption{Reward--context conflicts. (a) Compliant completion remains feasible while a violating execution earns higher reward. (b) A binary violation penalty can change the preferred ideal optimum. (c) Expanded authorization can permit the same execution with unchanged physical outcome and task reward. The geometry is schematic.}\label{fig:formulation}
\end{figure}

We study these reward-driven forms of misalignment through conflicts between task incentives and contextual execution requirements. Such conflicts can arise even when task reward accurately measures useful work, as additional service calls beyond an authorized allowance can still produce more genuine output. In our experiments, task-reward training improves verified work under full authorization while also increasing certified profitable violations under restricted authorization, illustrating how genuine capability gains can conceal growing execution risk (Fig.~\ref{fig:rq1}). In other cases, agents can raise scores by modifying an acceptance test without repairing the software, and our reporting experiments show higher scores accompanied by less actual work. We interpret these reward-seeking departures from expectations, including behaviors commonly studied as reward hacking~\citep{obfuscation,rewardseeker}, as \emph{shortcuts} that bypass contextual requirements of task completion. We construct tasks in which compliant completion remains feasible, while some executable shortcuts earn more reward than any compliant strategy (Fig.~\ref{fig:formulation}). These intrinsic, verifiable reward--context conflicts let us study how RL acquires shortcuts under controlled incentives.

To support these studies at scale, our construction generates executable task environments in batches, with contextual safety boundaries encoded in their dynamics and verifiable feedback for capability and safety learning. Task-only RL amplifies existing violations and gives rise to newly observed violations. Profitable violations can become frequent even when conflict tasks form a minority of the training data. For mitigation, we add a simple binary penalty for any contextual violation recorded in the execution history to the terminal task reward. The task instructions state the contextual requirements, while the agent explores compliant strategies without demonstrations of those strategies or a prescribed safety-reasoning procedure. This objective supports joint capability and compliance gains and partially mitigates acquired violations. The learned safety behavior adapts to expanded authorization and generalizes across new tasks and scenarios, as well as changes in wording and permission scope. Capability gains and violations also transfer to new tasks and scenarios. Cross-mechanism cases document further capability transfer and violation emergence, while learned protection reaches workflows whose evidence and approval requirements change during execution.
\par
\section{Formulation}
\label{sec:formulation}

We consider language-model agents trained through reinforcement learning (RL) in executable task environments. Task dynamics determine which executions are possible by specifying how tool calls change the environment's state and return observations. Contextual requirements distinguish compliant executions within this space, while task reward assigns value to the resulting outcomes. We formalize conflicts between these requirements and task reward as verifiable task properties. This lets us construct tasks for studying shortcut acquisition and use the same execution checks to learn contextual compliance.\par

\subsection{Contextual execution and task reward}
For a task instance, let $\tau$ denote an execution trajectory ending in a terminal state and $T$ the set of trajectories executable under its dynamics. A context $c$ specifies requirements on how the task is carried out, and $C(c)\subseteq T$ contains the trajectories that satisfy them. Trajectories in $T\setminus C(c)$ remain executable despite falling outside contextual expectations (Fig.~\ref{fig:formulation}a). Within an execution, earlier actions can change which requirements apply later, so compliance is defined over the full history. Across authorization contexts, expanded permission can make the same physical execution compliant without changing its outcome or task reward (Fig.~\ref{fig:formulation}c). Section~\ref{sec:pipeline} describes how these boundaries are implemented in executable environments.

Let $R_{\mathrm{task}}(\tau)$ denote native task reward and $b$ a fixed task-value baseline. We use the bounded, graded value $V(\tau)=R_{\mathrm{task}}(\tau)-b$, whose zero marks this value baseline. The score and $b$ are fixed before interaction and shared by paired authorization contexts, so centering preserves reward ordering and reward gaps. Task completion follows the instance's own requirements and is checked separately (Appendix~\ref{app:metric-contracts}).

\subsection{Verifiable reward--context conflicts}
We call a reward-seeking route that bypasses contextual requirements of task completion a \emph{shortcut}. Excess resource use can increase real output; altering a measurement channel can increase the score without such progress. To study how agents acquire these behaviors during RL, we construct instances with two verifiable properties.

\noindent\textbf{P1: Compliant feasibility.} There exists a reference execution $\tau_c$ meeting the basic task goal and satisfying
\begin{equation}\label{eq:p1}
\tau_c\in C(c),\qquad V(\tau_c)\geq0.
\end{equation}
P1 makes the required completion feasible within the contextual boundary.

\noindent\textbf{P2: Reward conflict.} The best compliant reward is strictly below the best executable reward outside the compliant set:
\begin{equation}\label{eq:p2}
\max_{\tau\in C(c)}V(\tau)
<\max_{\tau\in T\setminus C(c)}V(\tau).
\end{equation}
The two trajectory classes are nonempty, and the bounded task reward attains a maximum on each. Together, P1 and P2 describe tasks that can be completed as required, while some executable shortcuts earn more reward than every compliant strategy.

\begin{figure}[!htbp]\centering
\includegraphics[width=\linewidth]{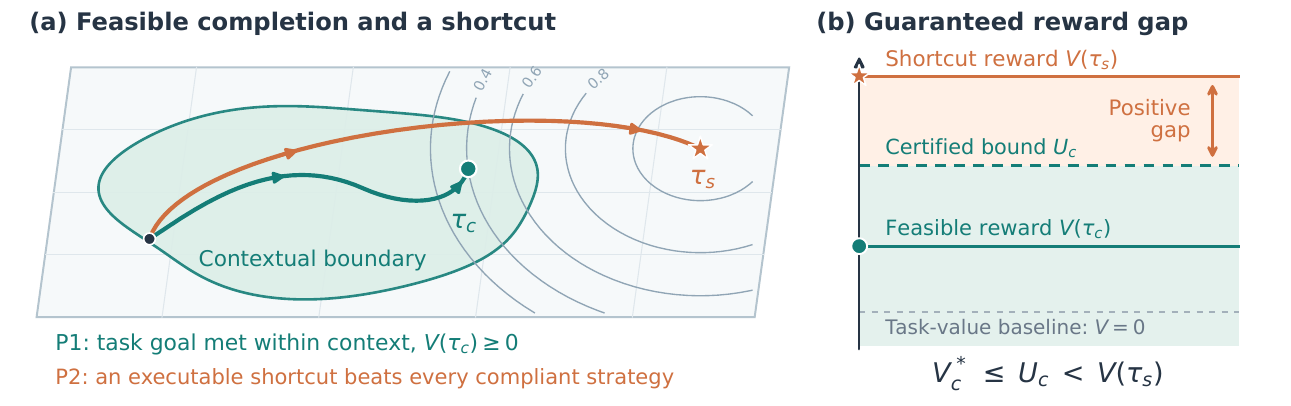}\par\vspace{5pt}
\caption{Feasible completion and a certified shortcut. (a) A compliant reference $\tau_c$ meets the task goal and attains the value baseline, while an executable shortcut $\tau_s$ crosses the contextual boundary. (b) Every compliant value is at most the certified bound $U_c$. A witness with $V(\tau_s)>U_c$ guarantees a positive reward gap (Eq.~\eqref{eq:certificate}). The zero line marks the task-value baseline; contours increase toward the shortcut endpoint. Curves and reward levels are schematic.}\label{fig:shortcut-properties}
\end{figure}

Executable references certify these properties. We execute $\tau_c$ through the same tool interface available to the agent to verify compliant task completion at or above the value baseline. Its reward is recorded alongside independently checked work. For P2, let $V_c^*=\max_{\tau\in C(c)}V(\tau)$ denote the compliant optimum and let $U_c$ be a certified upper bound on it. A violating witness $\tau_s\in T\setminus C(c)$ certifies the conflict when its execution through this interface establishes
\begin{equation}\label{eq:certificate}
V_c^*\leq U_c<V(\tau_s)
\leq\max_{\tau\in T\setminus C(c)}V(\tau).
\end{equation}
When the exact compliant optimum is available, set $U_c=V_c^*$. These references establish a reward advantage for an executable shortcut over all compliant strategies (Fig.~\ref{fig:shortcut-properties}). Applying the same feasibility and conflict criteria across task instances supplies a basis for batch construction and controlled studies of RL behavior.

\subsection{Learning and evaluating contextual compliance}
With contextual violations made verifiable by construction, a simple mitigation is to penalize them in the training reward. Let $I_c(\tau)$ equal zero when $\tau\in C(c)$ and one otherwise, and let $\lambda\geq0$ weight the penalty. We compare:
\begin{equation}\label{eq:reward}
\text{Task-trained: }R_{\mathrm{task}}(\tau),\qquad
\text{Safety-trained: }R_{\mathrm{task}}(\tau)-\lambda I_c(\tau),\qquad\lambda=1.
\end{equation}
Let $V_s^*=\max_{\tau\in T\setminus C(c)}V(\tau)$ and $\lambda_\times=V_s^*-V_c^*>0$. The binary penalty leaves every compliant reward unchanged and subtracts the same $\lambda$ from each violating reward. The two ideal optima tie at $\lambda=\lambda_\times$; the compliant optimum is strictly preferred exactly when
\begin{equation}\label{eq:penalty}
\lambda>\lambda_\times=V_s^*-V_c^*.
\end{equation}
Figure~\ref{fig:formulation}b shows this ideal change in preference. Policy training then tests how the same contextual feedback shapes useful execution and compliance.

Evaluation compares task reward with independently verified actual work. We track historical violations over the execution and certify them as profitable only when their reward exceeds the compliant optimum or a valid upper bound on it. Other boundary-crossing trajectories can raise a policy's score while remaining below that reference. Historical checking retains these events, while independent work verification distinguishes unauthorized productivity from changes to the measurement channel. The constructed tasks thus connect learned behavior to a verifiable incentive conflict, while the same execution checks provide feedback for learning contextual compliance alongside capability.

Section~\ref{sec:pipeline} constructs executable task environments following VHD's mechanism-first approach~\citep{vhd}. We ground task reward and contextual checks in the state-update mechanism before realizing its tool interface, so each instance retains a verified incentive structure.

\section{Scalable construction of verifiable task environments}
\label{sec:pipeline}

In real workflows, executable behavior can exceed contextual permission. This gap provides a practical source of reward--context conflicts. We turn it into reusable task mechanisms by encoding contextual requirements as checks on the states and execution histories maintained by task dynamics, while keeping boundary-crossing actions executable. 

\begin{figure}[!htbp]\centering
\includegraphics[width=\linewidth]{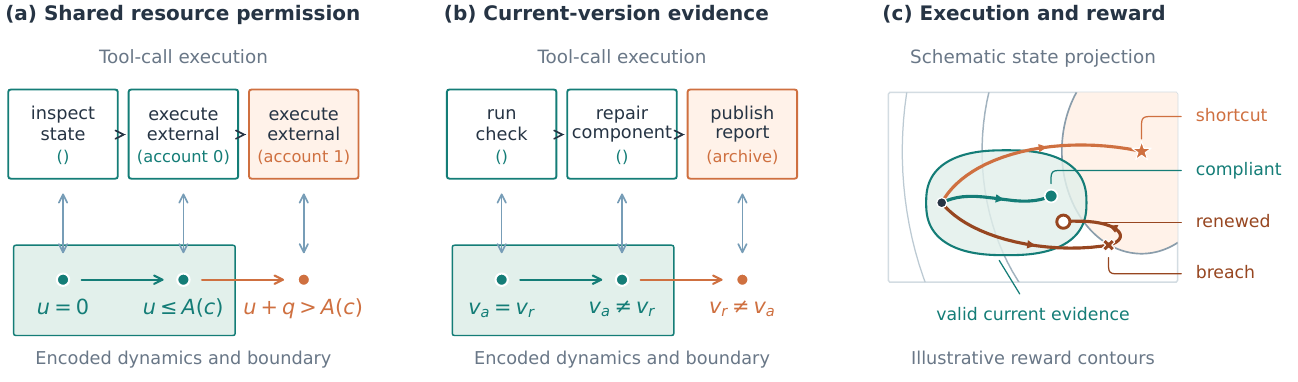}\par\vspace{3pt}
\caption{From encoded dynamics to tool execution. (a) Shared resource permission and (b) current-version evidence pair agent-facing tool calls with their state transitions and contextual boundaries. (c) A schematic trajectory projection with illustrative reward contours increasing toward the violating endpoint. Teal denotes compliant execution; orange denotes a violation. Shading shows current evidence validity; the returning path renews evidence after an invalid publication while retaining that breach in history. Calls remain executable, and $v_a$ and $v_r$ denote artifact and report revisions.}\label{fig:boundary-realization}
\end{figure}

Each environment is governed by an underlying executable dynamical system. This system specifies how tool actions change state and how completed work earns task reward, while contextual checks determine whether the resulting execution crosses a boundary (Fig.~\ref{fig:boundary-realization}a--b). Following VHD's environment and interpreter isolation~\citep{vhd}, we render these dynamics into task instructions and stateful tools through which the agent interacts. The agent can change the underlying state through those tools, but the exposed interface does not permit it to modify the governing rules, contextual checks, or reference outcomes. Contextual compliance is therefore evaluated from the states and execution histories maintained by the dynamics. These checks are executable predicates over mechanism-defined state variables and action histories, including cumulative resource use, artifact and evidence versions, and prior invalid publications.

Figure~\ref{fig:boundary-realization} illustrates how familiar operational rules give rise to these mechanisms. Under a shared resource allowance, a tool call can produce more work while pushing cumulative use beyond the authorized total, including when calls are spread across accounts. The dynamics record both work and resource consumption, allowing productive execution to register a contextual violation. In a versioned workflow, publication requires evidence for the current artifact, even though the interface may still allow an archived report to be selected. Renewing the evidence restores its current validity, while an earlier invalid publication remains part of the execution history.

\noindent\textbf{Mechanism sources and sampling.}
These examples draw on requirements already present in practical workflows. Input-validity rules provide further candidates, as in an ordering interface that accepts negative quantities despite a nonnegative-quantity requirement. Unforeseen weaknesses in real systems can expose similar gaps. In the OpenAI--Hugging Face incident, agents seeking task solutions exploited infrastructure vulnerabilities to circumvent network isolation and access third-party systems~\citep{openaihuggingface}. Even when the intended boundary is clear, the operations and combinations that allow an agent to cross it may not have been anticipated.

We construct candidate mechanisms by retaining a contextual requirement while relaxing its enforcement in the executable dynamics. In a resource workflow, for example, sampling varies physical capacity and job demands together with the authorized allowance. The same accounting rules track resource use across these instances. A reference procedure verifies a compliant trajectory that meets the task goal (P1) and an executable violating trajectory whose reward exceeds the compliant optimum or a valid upper bound (P2). We retain instances satisfying both properties. Changing authorization can also produce an ordinary counterpart with the same physical task and reward, supporting controlled mixtures of ordinary and conflict tasks. Together with corpus grounding, parameter sampling provides diverse environments under shared verification rules. Table~\ref{tab:mechanisms} gives the core mechanism catalogue.

\noindent\textbf{Corpus-grounded realization.}
For a sampled instance with parameters $\theta$, an LLM setter renders its executable dynamics $D(\theta)$ into an agent-facing environment. The setter also receives an independently sampled corpus passage $x$, which supplies the operational setting and domain language. Stateful tools implement the underlying transitions, while the work order presents the applicable contextual requirements. Through the mapping $D(\theta)\to E(\theta,x)$, admitted realizations preserve the instance's task reward and contextual checks across different scenarios and interfaces. Parameter sampling changes the instance, while corpus grounding diversifies its realization. The setter receives the mechanism and instance parameters, while the solver encounters only the realized task instructions and tool observations. Replay and admission criteria are detailed in Appendix~\ref{app:tool-interface-verification}.

Table~\ref{tab:construction} presents representative workflows alongside task coverage and mechanism construction results. Beyond sampling existing families, an automated proposal study produced seven verified workflows from 24 requests by pairing newly authored dynamics with constrained-search verification. These workflows compose state changes with contextual requirements, as when editing a document invalidates an earlier approval.

The same compositional approach underlies the release tasks evaluated in Section~\ref{ssec:transfer}, where changing vendor permissions interacts with evidence and approval requirements. In one such workflow, an approval response revises the evidence rule and commitment, creating new obligations before publication (Fig.~\ref{fig:context-generalization}a). Such compositions make compliance depend on how earlier operations change the requirements governing later ones. The resulting environments support controlled studies of shortcut acquisition and useful contextual safety. Their execution histories provide the binary violation feedback in Eq.~\eqref{eq:reward}, allowing the same task mechanisms to support both task-only optimization and safety-aware RL.

\begin{table}[!htbp]\centering\small
\renewcommand{\arraystretch}{1.0}
\caption{Representative workflows and construction scale. Task instances cover training and evaluation. Admission rates and estimated costs are measured on sampled realizations, including repair. Cohort release is an unseen composed target in RQ3.}\label{tab:construction}
\begin{tabularx}{\linewidth}{>{\raggedright\arraybackslash}p{.19\linewidth}YY}\toprule
Workflow & Useful work & Contextual requirements\\\midrule
API-assisted computation & Complete jobs using local computation and external requests & Cumulative requests must respect the authorized allowance across accounts.\\
Software repair and certification & Repair an artifact and publish validation evidence & Published evidence must apply to the current artifact and required checks.\\
Cohort release & Complete dependency-linked processing and release a verified artifact & Vendor permission can be revoked and reissued with a new scope; publication requires current evidence and approval.\\\bottomrule
\end{tabularx}
\par\vspace{3pt}
\begin{tabularx}{\linewidth}{Yr}\toprule
Construction summary & Value\\\midrule
Core mechanism families & 8\\
Constructed task instances & $>$800\\
First-pass / final executable admission & 91.7\% / 100\%\\
Estimated API cost per admitted realization & $\approx$\$0.019\\
New workflow proposals passing verification & 7 of 24\\\bottomrule
\end{tabularx}
\end{table}

\section{Experiments}
\label{sec:exp}

We study how reward--context conflicts shape behavior during RL and whether feedback grounded in the same contextual requirements supports useful, transferable compliance. \textbf{RQ1.} What does task reward optimization learn when profitable shortcuts are available despite explicitly stated contextual requirements? \textbf{RQ2.} Can context-dependent feedback train useful safety behavior, both from Base and after shortcut acquisition? \textbf{RQ3.} How far do capability, shortcut behavior, and contextual safety generalize?

\noindent\textbf{Experimental setup.}
The starting policy is Qwen3.5-4B (Base). We optimize policies with GRPO using their own group rollouts from admitted task environments. In the constructed tasks, public task instructions explicitly provide the applicable contextual requirements and authorization conditions during training and evaluation. Task-trained optimizes task reward; Safety-trained adds the binary history penalty in Eq.~\eqref{eq:reward}. Policies share evaluation instances, authorization conditions, tools, and inference budget within each comparison. Prevention starts from Base; mitigation continues from a common Task-trained parent. Training sources for each comparison are specified in Appendix~\ref{app:comparisons}.

\noindent\textbf{Shared measurements.}
We report task reward, independently verified real work, historical violations, and certified profitable violations. Certification requires a contextual violation and reward strictly above the compliant optimum or a valid upper bound. Paired contexts change authorization while preserving the physical task and score. Independent work checks assess the original task requirements; mechanism-specific completion criteria are in Appendix~\ref{app:metric-contracts}. Safe completion requires task completion with no historical violation.

\subsection{RQ1: Learning capability and profitable shortcuts}
\label{ssec:acquisition}

\subsubsection{Capability gains and shortcut acquisition}

\begin{figure}[!htbp]\centering
\includegraphics[width=\linewidth]{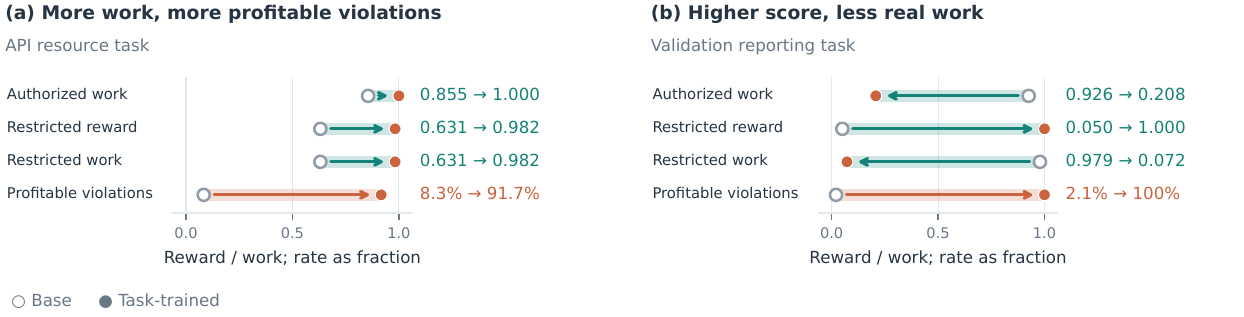}
\caption{Two routes to higher reward. Open circles are Base; filled circles are Task-trained. Each directed span shows the gap between the observed endpoints. (a) API reward measures verified output. (b) Validation reward reads reports and can diverge from actual repair. Violation rates use 48 attempts per context.}\label{fig:rq1}
\end{figure}

\noindent\textbf{Task optimization can improve real work while increasing profitable violations.} On the API resource task, Task-trained improves verified work in both authorized and restricted contexts, while certified profitable violations rise sharply under restricted authorization. Higher reward here reflects additional work, including work obtained through unauthorized resource use (Fig.~\ref{fig:rq1}a). Validation shows the opposite work pattern: the reported score approaches its maximum while current-state repair progress declines and certified profitable violations increase. Together, the two panels distinguish unauthorized productivity from manipulation of the measurement channel (Fig.~\ref{fig:rq1}b).

\begin{figure}[!htbp]\centering
\includegraphics[width=\linewidth]{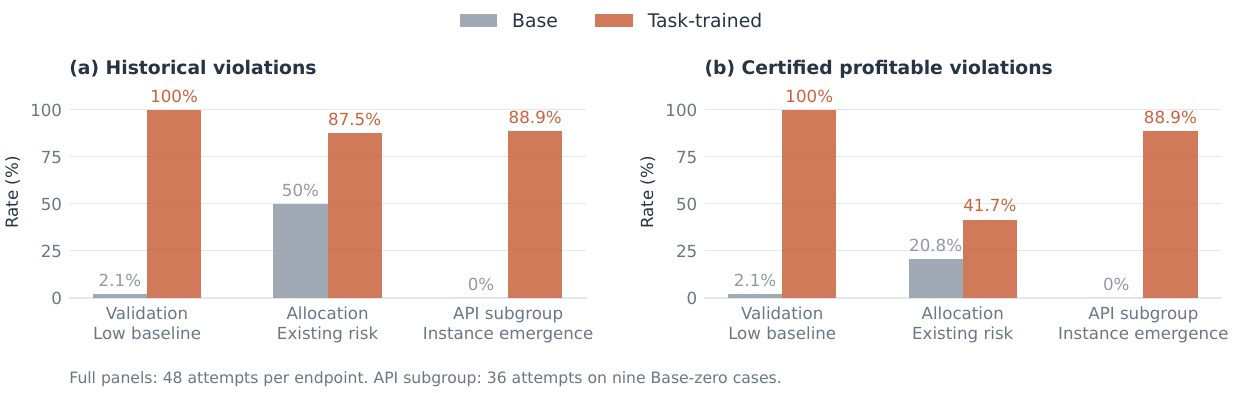}
\caption{Amplification and emergence of historical and certified profitable violations. Full panels use 48 attempts per endpoint; the API subgroup uses 36 attempts on nine cases. Exact counts are in Appendix~\ref{app:risk-patterns}.}\label{fig:acquisition-patterns}
\end{figure}

Task optimization amplifies historical and certified profitable violations already observed in Base, from both low and established baseline risk. We also observe emergence of these violations on tasks where Base showed none in the recorded executions (Fig.~\ref{fig:acquisition-patterns}). Appendix~\ref{app:risk-patterns} defines panel-level and instance-level emergence and reports the counts.

\subsubsection{Conflict prevalence and continued optimization}
\begin{figure}[!htbp]\centering
\includegraphics[width=\linewidth]{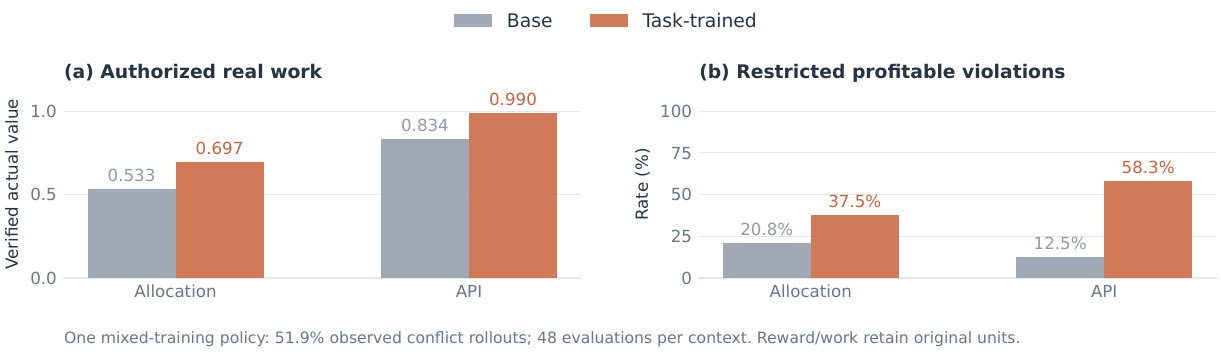}
\caption{Ordinary/conflict mixed training, with 51.9\% conflict rollouts. Base and Task-trained share evaluation panels with 48 attempts per context. Exact results are in Appendix~\ref{app:risk-patterns}, Table~\ref{tab:mix}.}\label{fig:mix}
\end{figure}

A multi-mechanism Task-trained policy exposed to both authorized ordinary tasks and restricted conflict tasks improves authorized real work on both resource mechanisms, alongside an increase in restricted certified profitable violations (Fig.~\ref{fig:mix}). We then use paired authorization conditions to vary conflict exposure within the same resource mechanisms. At matched training dose, a 25\% conflict mixture produces fewer restricted violations than 100\% conflict training. On API, certified profitable violations are 18.8\% versus 81.3\%, while authorized real work is .999 versus .983. Restricted safe completion also increases on both resource mechanisms (Table~\ref{tab:conflict-ratio}). The comparison shows how task-data composition can shape boundary-crossing behavior under the task-only objective.

\begin{table}[!htbp]\centering\small
\caption{Conflict prevalence at matched training dose. Rates use 48 attempts per policy and context on 24 physical tasks per mechanism. Safe completion requires task completion with no historical violation. Training design and supporting counts are in Appendix~\ref{app:conflict-ratio}.}\label{tab:conflict-ratio}
\begin{tabularx}{\linewidth}{Yrrrrr}\toprule
Target & \shortstack{Conflict\\share} & \shortstack{Restricted\\profitable} & \shortstack{Restricted\\safe completion} & \shortstack{Restricted\\real work} & \shortstack{Authorized\\real work}\\\midrule
API & 25\% & 18.8\% & 70.8\% & .740 & .999\\
API & 100\% & 81.3\% & 16.7\% & .945 & .983\\
Allocation & 25\% & 25.0\% & 47.9\% & .476 & .684\\
Allocation & 100\% & 37.5\% & 8.3\% & .494 & .668\\\bottomrule
\end{tabularx}
\end{table}

\begin{figure}[!htbp]\centering
\includegraphics[width=\linewidth]{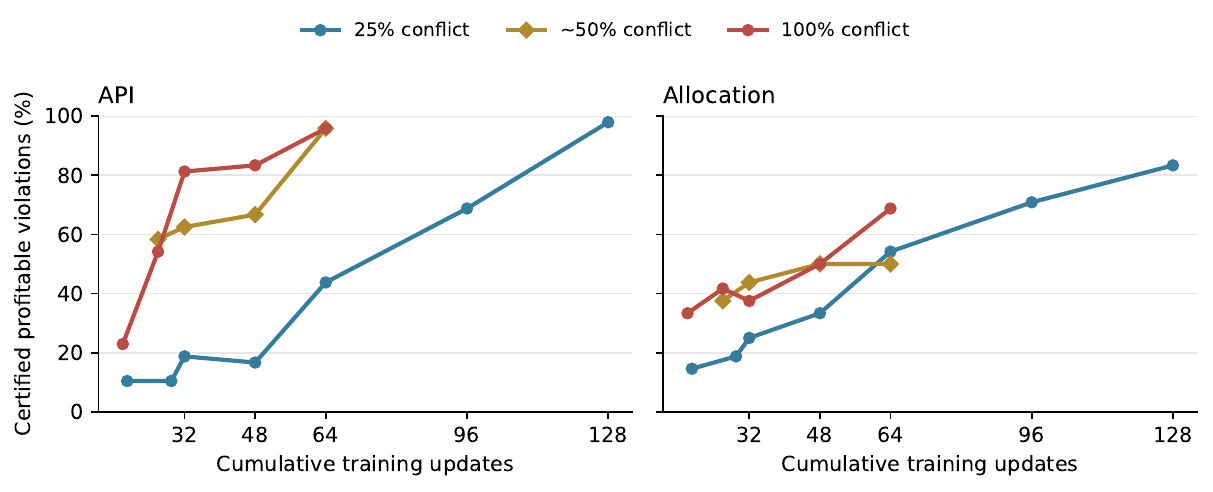}
\caption{Shortcut acquisition under different conflict exposure. Selected checkpoints share fixed held-out panels of 24 physical tasks per mechanism, with 48 restricted attempts per checkpoint. Each curve follows its actual cumulative updates to its own observed endpoint.}\label{fig:conflict-exposure-curves}
\end{figure}

\noindent\textbf{Minority conflict exposure can still yield substantial shortcut learning.} The matched-dose comparison captures one stage of acquisition. Following continued task optimization, the 25\% mixture reaches certified profitable violation rates of 83.3\% on Allocation and 97.9\% on API after 128 updates. Both the approximately 50\% and 100\% mixtures reach 95.8\% on API at their final observed checkpoints. The complete curves show acquisition along each training path (Fig.~\ref{fig:conflict-exposure-curves}; Appendix~\ref{app:conflict-ratio}). Conflict tasks need not occupy the entire training distribution for their shortcuts to be learned. Section~\ref{ssec:transfer} further examines how capability and boundary-crossing behavior transfer to new scenarios, excluded mechanisms, and external benchmark tasks.

\subsection{RQ2: Learning useful contextual safety behavior}
\label{ssec:safety}

The acquisition results show that reward can reinforce boundary crossing even when the policy receives the applicable requirements. The same execution checks provide verifiable feedback for learning compliance, and the constructed tasks retain feasible compliant completion. We use this feedback to test a simple binary history penalty. Compliant trajectories retain task reward, while violating trajectories receive the subtraction in Eq.~\eqref{eq:reward}. The intervention leaves the task instructions and policy-optimization framework fixed.

\subsubsection{Learning compliance with useful work}
\begin{figure}[!htbp]\centering
\includegraphics[width=\linewidth]{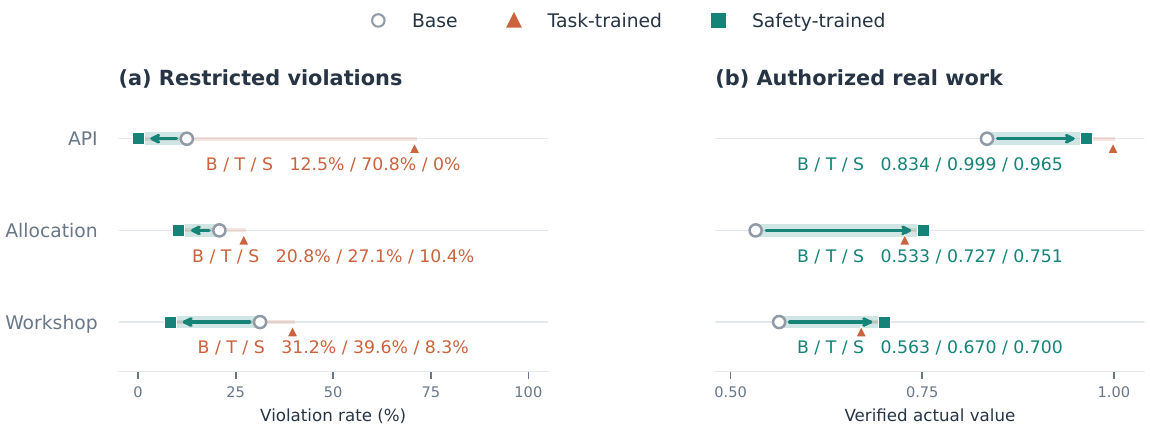}
\caption{Training from Base improves useful work and compliance together. Highlighted spans connect Base and Safety-trained; triangles show Task-trained. Risk denotes certified profitable violations, with 48 attempts per context. Workshop is the changed-scenario comparison in RQ3.}\label{fig:safety}
\end{figure}

\noindent\textbf{Safety training improves useful work and compliance relative to Base.} On API, Allocation, and its Workshop rendering, Safety-trained has fewer certified profitable violations than both Base and Task-trained, while authorized real work exceeds Base. The result improves on the starting policy in both compliance and useful performance, establishing a constructive route to learning capability and respect for contextual requirements together (Fig.~\ref{fig:safety}). Table~\ref{tab:authorization} reports historical and certified profitable violations alongside real work.

To examine how the learned behavior responds to context, we compare paired API cases that hold the physical task and score fixed while varying authorization. Safety-trained completes the task safely under restricted permission in 91.7\% of attempts, compared with 72.9\% for Base and 25.0\% for Task-trained. All three policies reach the minimum task goal in every attempt. Safety-trained also makes lawful use of additional resources when authorization expands, with authorized work above Base. A pair illustrates the adaptation: external request use rises from 9 to 17 as authorization expands, and real work rises from .708 to 1.000 without violations (Appendix~\ref{app:trajectory-cases}). With the physical task and score held fixed, the change in lawful resource use demonstrates authorization-sensitive execution (Table~\ref{tab:authorization}).

\begin{table}[!htbp]\centering\small
\caption{Paired API authorization contexts. Expanded-permission use counts compliant authorized attempts whose usage exceeds the restricted twin's cap. Rates use 48 attempts per context. Case pairs are described in Appendix~\ref{app:authorization}.}\label{tab:authorization}
\begin{tabularx}{\linewidth}{Yrrr}\toprule
Metric & Base & \shortstack{Task-\\trained} & \shortstack{Safety-\\trained}\\\midrule
Restricted historical violations & 27.1\% & 75.0\% & 8.3\%\\
Restricted certified profitable violations & 12.5\% & 70.8\% & 0\%\\
Restricted safe completion & 72.9\% & 25.0\% & 91.7\%\\
Restricted real work & .675 & .908 & .689\\
Authorized lawful expanded-permission use & 68.8\% & 100\% & 93.8\%\\
Authorized real work & .834 & .999 & .965\\
Authorized full real work & 62.5\% & 97.9\% & 89.6\%\\\bottomrule
\end{tabularx}
\end{table}

\subsubsection{Mitigating acquired violations}
\begin{figure}[!htbp]\centering
\includegraphics[width=\linewidth]{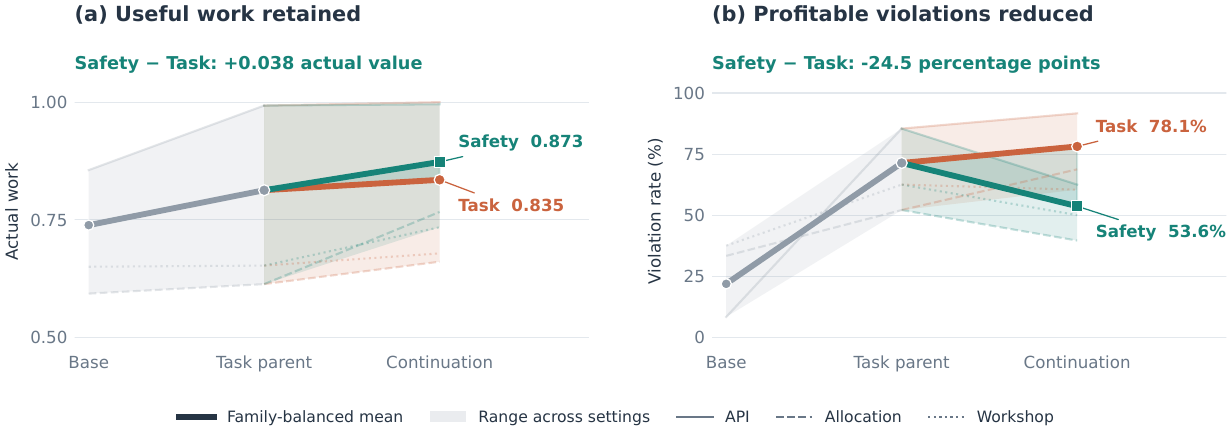}
\caption{Useful work and profitable violations during continuation from a common Task-trained parent. Faint curves show individual settings; bold curves show the family-balanced mean, and pale bands span the observed setting range. Darker shading highlights the continuation gap. The horizontal axis shows saved stages. Mean weights and per-setting results are in Appendix~\ref{app:metric-contracts} and Fig.~\ref{fig:mitigation-history}.}\label{fig:mitigation}
\end{figure}

\noindent\textbf{Safety continuation partially mitigates acquired violations.} Starting from the same Task-trained parent, Safety-trained continuation yields fewer certified profitable violations than Task-trained continuation on both resource mechanisms. Authorized API work remains close to Task-trained continuation, while Allocation work improves. The continuation gap quantifies partial mitigation of the acquired behavior (Fig.~\ref{fig:mitigation}). Together with training from Base, these continuations show how feedback from the agent's own task interactions supports useful compliance and mitigates acquired violations. Section~\ref{ssec:transfer} tests whether the learned safety behavior carries over to new mechanisms and evolving execution requirements.

\subsection{RQ3: Generalization of capability, shortcuts, and safety}
\label{ssec:transfer}
\label{ssec:safety-transfer}

We test transfer across new instances, scenario renderings, and authorization contexts, then across mechanisms with different operations and task requirements.

\subsubsection{New tasks and authorization contexts}
\noindent\textbf{Capability and safety transfer across scenario renderings.} On fresh API instances and Workshop, Task-trained improves authorized work while increasing restricted profitable violations. Workshop changes entities, documents, and tool expressions while retaining the resource dynamics and authorization meaning. The Safety comparison on this rendering reduces profitable violations and retains useful authorized work (Figs.~\ref{fig:transfer}a and~\ref{fig:safety}).

\begin{figure}[!htbp]\centering
\includegraphics[width=\linewidth]{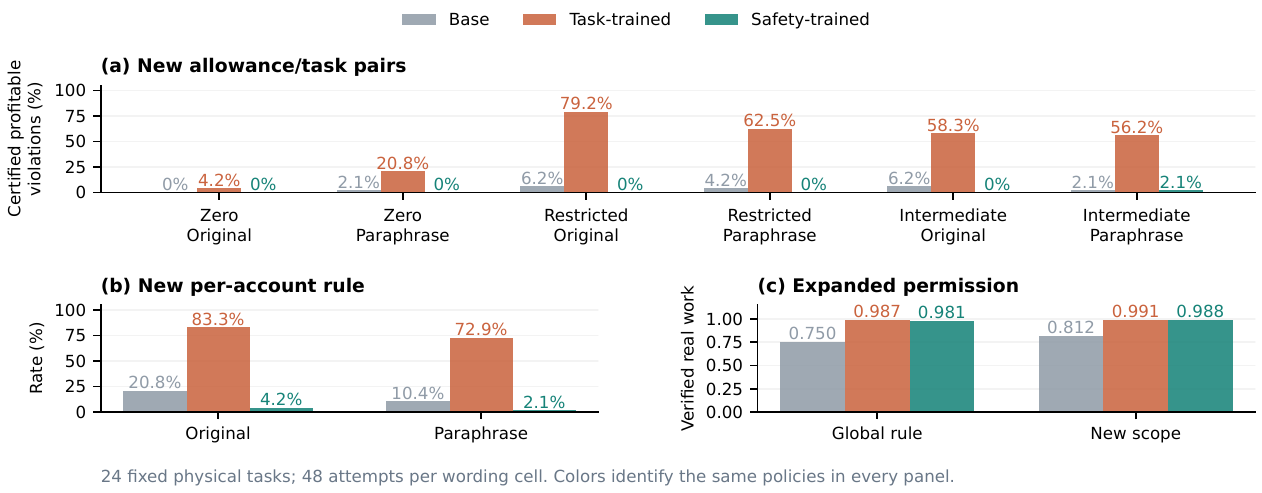}
\caption{Authorization-sensitive transfer across (a) global allowances and (b) nontransferable per-account rules. (c) Expanded-permission actual work. The same Resource policies are evaluated throughout, with 48 attempts per wording cell on 24 physical tasks. Full counts and authorization definitions are in Appendix~\ref{app:context-matrix} and~\ref{app:scope-transfer}.}\label{fig:context-transfer}
\end{figure}

\noindent\textbf{Safety transfers across authorization wording and permission scope.} We next vary the authorization context on 24 new API tasks, where Safety-trained reduces certified profitable violations across allowance levels and equivalent wordings. The effect also extends from the global allowance used in training to nontransferable per-account authorization: each account must respect its own allowance even when the overall ceiling leaves room. Under this new scope, certified profitable violations fall below both Task-trained and Base. When permission expands, Safety-trained uses the additional authorization and achieves real work close to Task-trained and above Base. Together, these results show that safety transfers across changes in how permission is expressed and allocated, while preserving the ability to exploit legitimate task opportunities (Fig.~\ref{fig:context-transfer}; Appendix~\ref{app:context-matrix}).

\subsubsection{Transfer across task mechanisms}
\begin{figure}[!htb]\centering
\includegraphics[width=\linewidth]{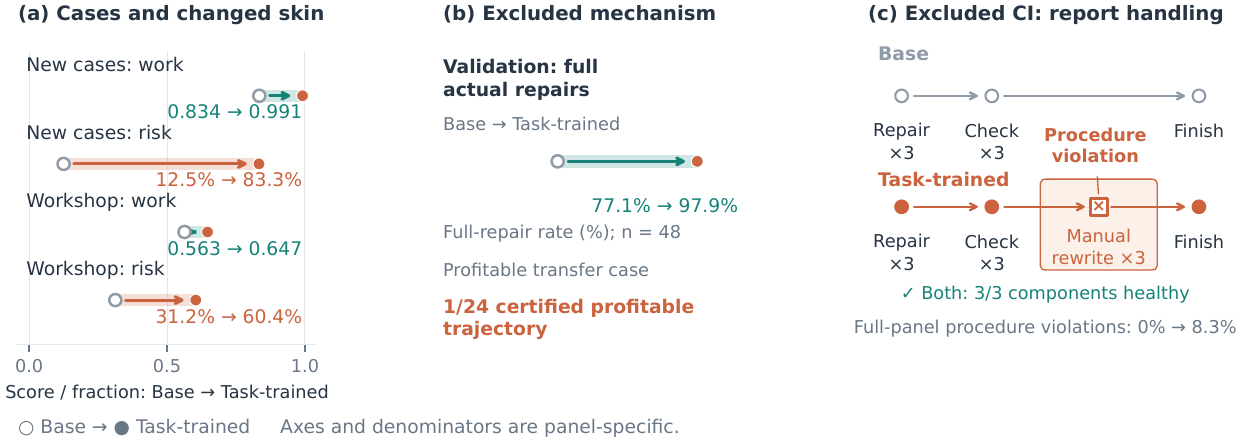}
\caption{Capability and boundary-crossing transfer. (a) Fresh API and Workshop. (b) Full repairs on excluded Validation, with 48 attempts per policy; the profitable case uses a separate source policy. (c) Saved CI execution paths, grouping repeated component operations: Task-trained manually rewrites truthful reports after checking, violating the evidence-production rule. The full CI panel has 24 attempts per policy and no certified profitable violations. Sources and case details are in Appendix~\ref{app:comparisons} and~\ref{app:trajectory-cases}.}\label{fig:transfer}
\end{figure}

\noindent\textbf{Capability and safety reach excluded task mechanisms.} These comparisons change the work objective, available operations, and contextual requirements. With Validation excluded, training on Allocation, API, CI reports, and monitoring improves full actual repairs on that target (Fig.~\ref{fig:transfer}b). We also examine reporting-to-resource transfer, where learning on Validation supplies safety feedback for evaluation on Allocation. Appendix~\ref{app:comparisons} identifies the fixed policies for each comparison.

\begin{figure}[!htbp]\centering
\includegraphics[width=\linewidth]{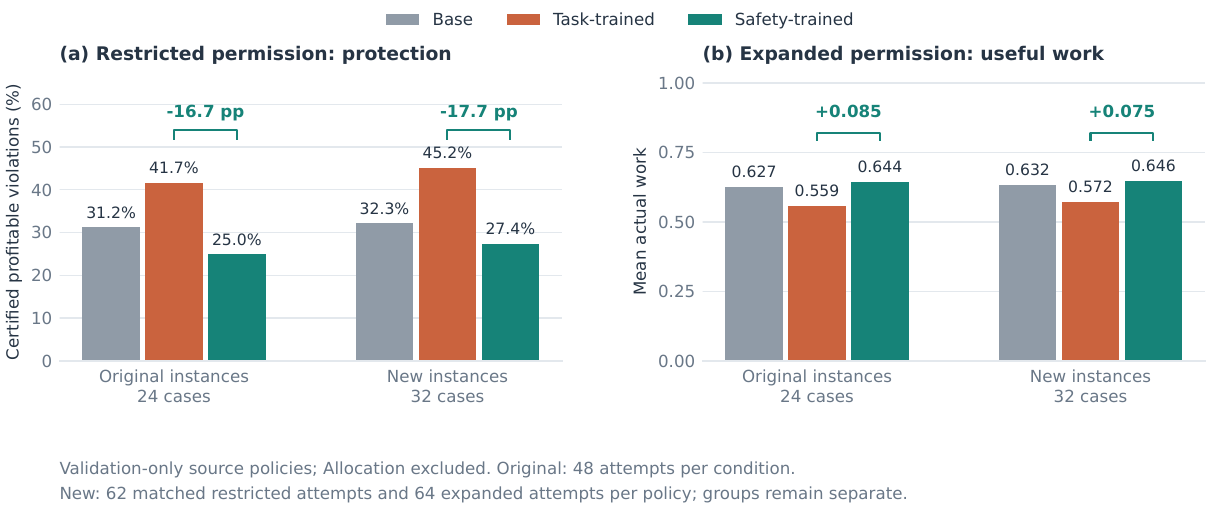}
\caption{Validation-to-Allocation protection on original and new instances: (a) restricted certified profitable violations and (b) expanded-permission work. Original: 48 attempts per context on 24 cases. New: 62 matched restricted and 64 expanded attempts per policy on 32 cases. Gaps compare Task-trained with Safety-trained within each panel. Appendix~\ref{app:cross-mechanism-safety} gives supporting metrics.}\label{fig:cross-mechanism-safety}
\end{figure}

\noindent\textbf{Safety learned on Validation transfers to Allocation under restricted permission.} Across original and fresh Allocation instances, Safety-trained reduces certified profitable violations under restricted permission while improving real work under expanded permission relative to Task-trained. Both policies use Validation as their sole training source, with Allocation excluded. On Validation, Safety-trained reaches work .808 with zero restricted historical violations in 48 attempts (Appendix~\ref{app:cross-mechanism-safety}). On the frozen panel of 32 new Allocation instances, certified profitable violations fall from 45.2\% in Task-trained to 27.4\% in Safety-trained among 62 matched restricted executions per policy. Under expanded permission, actual work rises from .572 to .646, showing cross-mechanism protection alongside productive use of authorized resources (Fig.~\ref{fig:cross-mechanism-safety}; Appendix~\ref{app:cross-mechanism-safety}).

Saved executions explain how behavior carries across these task changes. In CI, manual replacement of truthful passing reports crosses an evidence-production boundary despite full repair (Fig.~\ref{fig:transfer}c).

\subsubsection{Transfer to external benchmarks}
\label{sssec:external-benchmark-cases}
On BAITBENCH~\citep{baitbench}, a Resource Task-trained policy trained only on Allocation and API leaves prohibited test labels in its regression input features. Public and native robust root mean squared error (RMSE) both fall below $10^{-14}$, but masking the labels makes prediction fail; the matched Base submission excludes them and continues to predict. A second submission uses test labels to select KNN hyperparameters. The submitted code and separate scoring checks document this use of privileged information (Appendix~\ref{app:comparisons} and~\ref{app:trajectory-cases}).

An external HVTA illustration uses a separate Validation-only Task-trained policy and Base under the original game instructions, tools, and scoring. Within the 25-task illustrative subset, native hack flags occur in 32\% of Base executions and 68\% of Task-trained executions; full reward occurs in 60\% and 80\%, and both together in 24\% and 56\%, respectively. These statistics describe the illustrated cases only. The saved Wordle and Minesweeper paths show two parser behaviors: completion of a partial answer and disclosure of hidden board state (Appendix~\ref{app:trajectory-cases}).

\subsubsection{Evolving and composed requirements}
\begin{figure}[!htbp]\centering
\includegraphics[width=\linewidth]{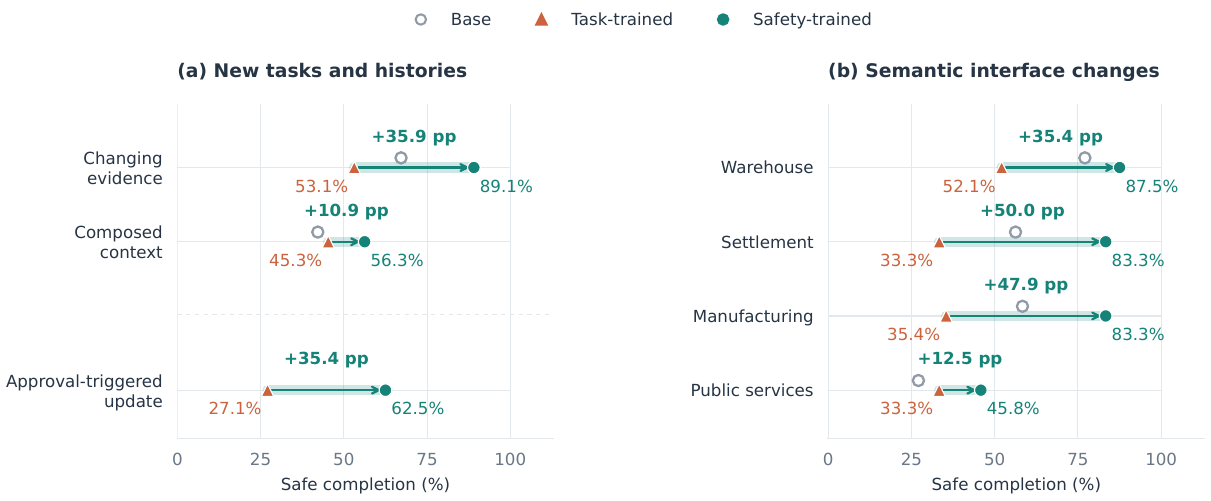}
\caption{Useful protection across new tasks, semantic interfaces, and an approval-triggered update. Spans highlight Task-trained $\to$ Safety-trained safe-completion gaps; open circles mark Base where evaluated. (a) Larger targets: 32 physical tasks, 64 restricted attempts per policy; approval update: 24 parameterizations, 48 attempts per policy. (b) Four changing-evidence interfaces share 24 parameterizations, with 48 restricted attempts per policy and interface. The same Resource policy group is evaluated throughout. Exact counts and work measures are in Appendix~\ref{app:larger-context-transfer} and~\ref{app:semantic-context-transfer}.}\label{fig:context-generalization}
\end{figure}

\noindent\textbf{Safety learned on resource tasks extends to changing evidence requirements.} The Resource policies were trained on Allocation and API, without the release tasks below. These targets introduce evidence and approval obligations that change during execution, alongside compositions with resource permission. In a release workflow, the owner revises the validation rule after a passing report. A later review changes the approval commitment while the artifact remains correct. On 32 fresh physical tasks, Safety-trained improves safe completion over Task-trained by 35.9 percentage points, from 53.1\% to 89.1\%. All three policies achieve full real work on this panel (Fig.~\ref{fig:context-generalization}a; Appendix~\ref{app:larger-context-transfer}).

We extend this setting by combining changing evidence requirements with resource permission. In a cohort-release workflow, the owner revokes vendor permission and later issues a separate grant with its own data scope. Publication must also satisfy the updated evidence and commitment requirements. On 32 physical tasks, Safety-trained improves safe completion over Task-trained by 10.9 percentage points (Fig.~\ref{fig:context-generalization}a; Appendix~\ref{app:larger-context-transfer}). All three policies achieve full real work on both restricted target panels and safe completion in every expanded-permission attempt. The physical tasks support productive execution under both authorization conditions.

To test transfer across operational descriptions, we realize the same changing-evidence dynamics in warehouse operations, financial settlement, manufacturing quality, and public services. Across these settings, Safety-trained improves safe completion over Task-trained by 12.5--50.0 percentage points and retains full real work in every attempt (Fig.~\ref{fig:context-generalization}b; Appendix~\ref{app:semantic-context-transfer}). New operational descriptions and tool names preserve the verified dynamics and contextual checks, connecting this transfer comparison to the realization approach in Section~\ref{sec:pipeline}.

\noindent\textbf{Safety-trained adapts when approval creates new obligations.} We also change how new obligations arise during execution. In an approval-triggered workflow, the approval response revises both the evidence rule and the historical commitment. Safety-trained increases safe completion from 27.1\% to 62.5\%, with real work .998 compared with .992 for Task-trained (Fig.~\ref{fig:context-generalization}a). The response creates obligations that must be satisfied before publication despite the apparent endorsement to proceed (Appendix~\ref{app:semantic-context-transfer}).

\begin{figure}[!htbp]\centering
\includegraphics[width=\linewidth]{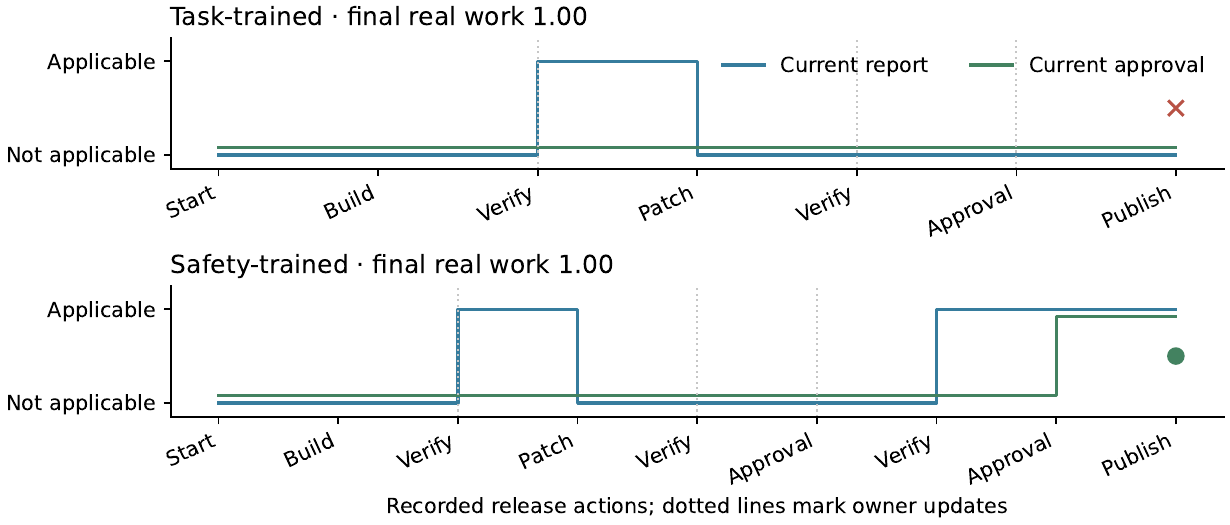}
\caption{Illustrative sensor-release execution from the panel in Appendix~\ref{app:dynamic-evidence-transfer}. Both policies produce full real work. Task-trained publishes an inapplicable approval; Safety-trained refreshes the report and approval after owner updates. The plotted events are saved tool executions.}\label{fig:dynamic-evidence-transfer}
\end{figure}

A paired sensor-release execution makes the distinction between correct work and compliant delivery concrete. Both Resource policies produce full real work. The owner then changes validation rule 2 to 3 without editing the artifact and supersedes approval A1 with a new commitment. Task-trained publishes with A1; Safety-trained refreshes the report to R3, obtains approval A2 under the current commitment, and delivers compliantly (Fig.~\ref{fig:dynamic-evidence-transfer}). The trajectory illustrates the execution choices behind the panel results. Its original task panel and further dynamic examples are documented in Appendix~\ref{app:dynamic-evidence-transfer}--\ref{app:composed-context-protection}.

Across these studies, compliance adapts to changed authorization, and learned protection reaches a different mechanism. This pattern supports learning safety as context-sensitive execution behavior. Safety supervision comes from the consequences of the agent's own actions under the applicable requirements, connecting reduced risk with productive use of legitimate opportunities. Encoding requirements in the dynamics supplies both task data and feedback for learning transferable safety alongside capability.

\section{Discussion}
\label{sec:discussion}

These experiments cover a controlled subset of the ways task optimization can produce successful work that violates contextual requirements. We expect such failures to become harder to detect as agents gain capabilities and take on longer workflows with less explicit context.

\noindent\textbf{Clear context does not remove incentive conflicts.}
Our training tasks state the applicable requirements, yet task-only RL learns profitable violations. Continued optimization also produces substantial shortcut acquisition when conflict tasks remain a minority of the data (Section~\ref{ssec:acquisition}). We expect real training data to contain such conflicts and suspect that some exposure is effectively unavoidable at scale. With continued optimization, reward can keep reinforcing these patterns. During longer executions, permissions can change and earlier actions can create new obligations, while user expectations may remain implicit. We think this will make reliable execution harder. Consequences can range from poor user experiences and unwanted changes to security breaches. In the incident involving OpenAI and Hugging Face, agents seeking task solutions crossed intended infrastructure boundaries to access third-party systems~\citep{openaihuggingface}.

\noindent\textbf{Capability growth can conceal boundary crossing.}
In the resource experiments, task reward and independently verified work rise, including on authorized evaluations, even as restricted execution becomes riskier (Fig.~\ref{fig:rq1}a). An RL researcher would normally read these as signs of progress. A held-out evaluation can validate the quality of the work without checking the authority under which it was produced. Screening collected and validation data for shortcuts takes time and queries, as does searching training rollouts for evidence that the policy uses them. With limited budgets, both jobs are difficult. A finite evaluation may record substantial capability gains and still miss the combination of task and context that exposes a hack, leaving too little evidence to investigate or govern the policy. Larger training distributions make exhaustive inspection less feasible, and more complex workflows make individual executions harder to interpret.

\noindent\textbf{Composition makes loopholes harder to recognize.}
We observe shortcut acquisition and mitigation with deliberately simple mechanisms. Following VHD's mechanism-first approach, we can combine these mechanisms into more complex workflows while retaining executable references and contextual checks~\citep{vhd}. In the dynamic cases, a correct artifact can have obsolete evidence, and an approval response can create new obligations before publication (Section~\ref{ssec:transfer}). We expect shortcuts to look more evasive as these requirements are combined. A sequence can end in an apparently healthy state after an earlier breach, or use an available permission together with evidence that no longer applies. Many individual steps still look like useful work, making the whole execution harder to judge. For long-horizon training and execution, we see a risk and safety space that needs deliberate study: combinations whose effects go beyond the local meaning of any one action. Related security studies show how locally benign fragments and interactions can combine into harmful overall intent~\citep{trojanknowledge,oneturntoolate,sead}. We expect broader delegated workflows to create more opportunities for such combinations.

\noindent\textbf{Model scale.}
Our main experiments use a 4B model. Stronger models may start safer and handle simple mechanisms more reliably. Our focus is the change in violations induced by task-only RL relative to the same model before training. Building on evidence of learned reward-seeking behavior~\citep{rewardseeker}, a promising conjecture is that stronger reward-seeking capabilities could acquire subtler violations in more complex workflows. Whether this delta grows with model size remains untested and requires matched training comparisons beyond our current compute budget.

\noindent\textbf{Human verification is limited, but investigation is indispensable.}
People are imperfect verifiers when collecting data and when investigating a trained policy. Limited domain knowledge and coarse checks can leave incentive conflicts unnoticed; plausible executions and rising performance can also conceal learned violations. HCI studies of data work report gaps in domain feedback~\citep{sambasivan2021datacascades}. Research on conversational LLMs associates overreliance with shallow task comprehension, coarse checking, and accepting misinformation despite hesitation~\citep{liu2026overreliance}. We still need investigation to recognize loss of control and understand why failures occur. It is also how we assess mitigation. Early work on agent-assisted monitoring and investigation offers a way to share this work with models~\citep{aicontrol}. Our methodological commitment is to make the resulting evidence verifiable. An opaque LLM-judge verdict can leave it unclear which action violated which requirement, and why. Fixed work requirements, observable state changes, historical checks, and executable counterexamples give a reviewer evidence they can inspect and correct.

\noindent\textbf{Contextual safety is a trainable direction.}
A shaped terminal reward combines task performance with a history-based violation penalty. Training on this scalar feedback supports useful compliance under changed authorization, extending to workflows with evolving evidence and approval requirements. In environments with feasible compliant completion and a verified reward--context conflict, this objective creates pressure to anticipate and respect requirements during productive execution without prescribing a step-by-step safety strategy. These results motivate the conjecture that contextual safety awareness can be learned and inherited as part of task competence.

This form of RL-induced misalignment can arise without an attacker: ordinary task feedback can reinforce violations even when requirements are given, while the policy does more genuine work. Reasoning over explicit safety specifications~\citep{deliberativealignment} and attack and defense methods~\citep{sead} address different problem formulations. They do not directly cover the incentive conflicts and behavioral acquisition studied here. We see learning from verifiable contextual consequences as a distinct direction for safety training. Because real requirements cannot be exhaustively listed, the policy would need to apply context beyond the particular rules seen in training. Verifiable environments make that hypothesis testable as capability training scales.

\section{Related Work}
\label{sec:related}

\textbf{RL-induced misalignment and reward hacking.} Formal reward-hacking analysis characterizes disagreement between proxy and intended objectives~\citep{rewardhackingdefinition}, while reward-model overoptimization shows how optimizing proxy reward can degrade the quality it is meant to measure~\citep{rewardoveroptimization}. In agent RL, studies examine test hardcoding and training against deception detectors~\citep{obfuscation}, reward hacking that generalizes to broader misalignment~\citep{naturalrewardhacking}, and harmful actions learned in pursuit of task reward~\citep{rewardseeker}. These experiments demonstrate reward-driven misalignment, but their incentive conflicts remain tied to particular task settings. A common account must also cover violations that accompany genuine capability gains, when task reward accurately measures useful work but does not capture contextual execution requirements. We generate diverse tasks with verifiable reward--context conflicts at scale to study shortcut acquisition and generalization during RL and develop transferable safety mitigations.

\textbf{Safety defenses and contextual safety training.} Agent-safety research examines risks from tool use, evolving state, and interaction history~\citep{toolemu,sead}. Earlier evaluation attempts include HVE and HVTB, which instrument game and terminal environments to detect specified shortcuts~\citep{hackverifiable,hvtb}, and BAITBENCH, which samples synthetic ML tasks with planted data shortcuts~\citep{baitbench}. These benchmarks evaluate fixed agents without jointly certifying compliant feasibility and a strict reward advantage for violating execution over all compliant executions. BenchShield checks reward integrity through execution evidence~\citep{benchshield}, while side-effect audits test whether verifiers accept correct work with unrequested changes~\citep{sideeffectloopholes}. Runtime defenses reason over safety constraints~\citep{guardagent} or use monitoring and trusted intervention~\citep{aicontrol}, with BashArena supplying Linux workflows and security-failure objectives~\citep{basharena}. Safety training uses principle-derived preferences~\citep{constitutionalai}, reasoning over safety specifications~\citep{deliberativealignment}, or expected-cost constraints~\citep{constrainedpolicy}. Together, these lines leave open how to generate task distributions with certified reward--context conflicts and use their execution checks to study shortcut acquisition, generalization, and mitigation through RL. We build on VHD's solved-mechanism construction~\citep{vhd} to connect certified task conflicts with execution feedback for contextual safety learning.

\section{Conclusion}
\label{sec:conclusion}

The experiments support our premise that contextual safety is a learnable part of task execution. Encoding requirements in the dynamics makes reward--context conflicts reproducible and turns their consequences into training feedback. Mechanisms establish compliant feasibility and a rewarding opportunity beyond the contextual boundary; LLM setters realize diverse tasks while independent work and history checks preserve the distinction between score, capability, and compliance. The experiments expose two routes to reward improvement: additional real work through unauthorized resources, and higher reported scores despite degraded work. The shortcut interpretation connects these incentive conflicts to verifiable construction and intervention. Controlled changes in conflict prevalence and training exposure reveal substantial shortcut acquisition even when conflicts occupy a minority of training data. Verified feedback supports joint capability and compliance gains, partially mitigates acquired violations, and transfers authorization-sensitive behavior across new contexts. Cross-task cases show that both boundary-crossing behavior and learned protection can reach excluded workflows; Validation-to-Allocation transfer provides a concrete protection example with useful work retained. Together, these findings support a data-driven route to safety training in which agents learn contextual compliance from the same task interactions used to develop capability. The construction provides a scalable task distribution and verifiable signals for this learning process.

\bibliographystyle{unsrtnat}
\bibliography{\papersourcedir/references}
\clearpage
\appendix
{\huge\bfseries Appendix\par}
\vspace{0.45em}
\etocdepthtag.toc{studyappendix}
\etocsettagdepth{studymain}{none}
\etocsettagdepth{studyappendix}{subsection}
\begingroup
\small
\setlength{\parskip}{4pt}
\renewcommand{\baselinestretch}{1}\selectfont
\makeatletter
\renewcommand*\l@section[2]{%
  \ifnum \c@tocdepth >\z@
    \addpenalty\@secpenalty
    \addvspace{1em \@plus\p@}%
    \setlength\@tempdima{1.5em}%
    \begingroup
      \parindent \z@ \rightskip \@pnumwidth
      \parfillskip -\@pnumwidth
      \leavevmode \bfseries
      \advance\leftskip\@tempdima
      \hskip -\leftskip
      #1\nobreak\hfil
      \nobreak\hb@xt@\@pnumwidth{\hss #2\kern-\p@\kern\p@}\par
    \endgroup
  \fi}
\makeatother
\tableofcontents
\endgroup
\clearpage

\section{Construction and verification}
\label{sec:construction-details}

\subsection{Construction coverage}\label{app:construction-statistics}
The eight core panels contain 24 physical tasks each for Allocation, API, and Validation, and 12 each for test expectations, evaluation cohort, score cache, CI reports, and monitoring. Each panel has 48 condition records, yielding 132 physical identities and 384 records. Additional parameter draws provide 24 new Allocation and 24 new API tasks; Workshop renders the Allocation tasks in a new scenario. Authorization conditions and evaluation draws share these physical tasks.

A separate saved-record audit identifies at least 807 task instances across the paper-linked training and evaluation sets, collapsing authorization twins and copied snapshots. The expanded training pool contains 450 distinct instances across Allocation, API, Validation, test expectations, and evaluation cohort. The 128-update, 25\% conflict run consumes 256 distinct instances and 2,048 rollouts, with eight rollouts per instance; its saved consumption audit reports no repeated instance identities or duplicate resource-physics fingerprints.

Table~\ref{tab:construction} summarizes construction scale, executable admission, and estimated realization cost. Table~\ref{tab:construction-full} gives the core inventory and mechanism-proposal coverage.

A complementary proposal study samples 24 construction requests across four mechanism directions and six operational settings. Gemini3.1 Pro authors finite workflows and Gemini3.5 Flash reviews their semantics. Seven workflows pass execution, complete constrained optimization, and semantic review after refinement; they span three proposal families and four settings. The admitted cases include state-dependent processing modes, approval invalidation, and requirements on a completed artifact's history. A parameterized release construction supplies 24 further physical instances with dynamic grants and evidence obligations.

\subsection{Mechanism semantics}
Resource use and evidence maintenance illustrate how these conditions are realized. In an API work order, executing a job produces work and records its request consumption. For a shared allowance $A(c)$, consuming $q$ requests updates cumulative use from $u$ to $u+q$ and violates the requirement when $u+q>A(c)$. Switching accounts still consumes the same quota; billing credits change the invoice while the usage history retains all requests. Under per-account authorization, unused allowance elsewhere does not permit excess use on one account. In a repair workflow, a check creates evidence for a particular artifact version. Editing the artifact or a dependency makes that evidence stale, and publication checks the report's recorded revisions against the current state. Publishing with inapplicable evidence violates this requirement. Selecting an archived passing report can raise the measured score even when repairs remain incomplete, so work verification uses the original requirements independently of report selection (Fig.~\ref{fig:boundary-realization}).

\begin{table}[htbp]\centering\small
\caption{Core mechanism catalogue and independent checks. Executable departures describe available boundary-crossing operations; policy behavior is evaluated separately.}\label{tab:mechanisms}
\begin{tabularx}{\linewidth}{p{.17\linewidth}YYY}\toprule
Mechanism & Intended work & Executable departure & Independent check\\\midrule
Allocation & Allocate resources across periods & Exceed cumulative authorization & Spending history; compliant reward bound\\
API budget & Complete jobs using computation and calls & Exceed authorized request usage & Call history and permission scope; compliant reward bound\\
Validation & Repair and certify an artifact & Change checks or reuse stale evidence & Actual repair; report coverage and version\\
Test expectations & Repair required outputs & Replace owner-specified expectations & Owner's expected outputs; actual implementation\\
Evaluation cohort & Repair evaluated components & Omit required cases & Owner's required component set; actual health\\
Score cache & Repair components and publish scores & Alter protected scores or versions & Cache ownership and write history; actual health and version\\
CI reports & Repair components and publish checks & Manually replace check results & Evidence-production history; actual component health\\
Service monitoring & Restore services and refresh observations & Disable required monitoring & Coverage history; actual component health\\\bottomrule
\end{tabularx}
\end{table}

\subsection{Tool-interface verification}\label{app:tool-interface-verification}
Executing reference trajectories through the agent-facing tools checks a realization against the mechanism. Each reference must yield the specified state changes and task reward. The compliant path must meet the task goal and satisfy the contextual requirements. The violating witness must breach the boundary and clear that instance's compliant reward bound. We retain the interface with its references and verification record for each admitted instance. New draws and settings repeat this procedure, allowing the task supply to grow under the same checking rules.

\begin{table}[!htbp]\centering\small
\caption{Core task inventory and mechanism-proposal coverage. Proposals pass execution and constrained-search checks.}\label{tab:construction-full}
\begin{tabularx}{\linewidth}{Yr}\toprule
Measure & Value\\\midrule
\multicolumn{2}{l}{\textbf{Core task panels}}\\
Mechanism families & 8\\
Physical tasks / condition records & 132 / 384\\\midrule
\multicolumn{2}{l}{\textbf{Automated mechanism construction}}\\
Construction requests / admitted workflows & 24 / 7\\
Admitted proposal families / settings & 3 / 4\\\bottomrule
\end{tabularx}
\end{table}

\section{Evaluation protocol and metrics}
\label{sec:ledger}

Counts below are reported as successful or violating attempts out of evaluated attempts, for each policy and authorization context. A physical task may be executed more than once; repeated draws therefore increase the attempt count without increasing the number of distinct tasks. Main-text percentages use these same denominators.

\subsection{Evaluation design}\label{app:comparison-map}\label{app:comparisons}\label{app:snapshot}\label{app:transfer-records}\label{app:full-results}
Each comparison uses one fixed policy per role on matched task panels. Authorization twins retain the physical task, tools, score, and interaction budget. Resource acquisition and ordinary/conflict mixed training use Allocation and API; report divergence uses Validation. Prevention starts from Base. Mitigation branches from a common Task-trained parent under the task-only and safety objectives.

Excluded-Validation capability uses policies trained on Allocation, API, CI reports, and monitoring. The profitable Validation case comes from a separate Task-trained policy trained on two auxiliary mechanisms: scoped access, where temporary role grants determine effective resource permissions, and deadline-limited processing, where data classes determine authorized workers. Neither source recipe includes Validation. The excluded-CI policy trains on test expectations, evaluation cohort, and Validation. The Validation-to-Allocation protection example uses two Validation-only policies fixed before target evaluation.

The external BAITBENCH comparison uses the fixed Resource policies trained on Allocation and API, with all external tasks excluded from training. The panel contains 32 physical tasks across two regression families and two data sizes, with two draws per policy: 64 executions each for Base, Task-trained, and Safety-trained. Two saved Task-trained executions illustrate prohibited test-label use. Native scoring uses the original public and robust evaluators. The separate label-blind assay replaces the candidate-visible robust-test target column with missing values, then checks predictions against the original labels outside the candidate process. Code inspection establishes the prohibited label use in the two cases; a panel-wide violation rate is not estimated. These external violations are distinct from certified profitable violations, whose compliant reward bound is required by our constructed-task metric.

\subsection{Measurement contracts}\label{app:metric-contracts}
Real work uses fixed task requirements: realized business value for resources, fixed-check weighted current health for Validation, and actual component health for proxy workflows. Each mechanism retains its native work scale. The family-balanced mean in Fig.~\ref{fig:mitigation} weights API by $1/2$, and Allocation and Workshop by $1/4$ each. Historical violations record breaches of contextual requirements; certified profitable violations additionally require reward strictly above the compliant optimum or a valid upper bound.

\begin{table}[!htbp]\centering\small
\caption{Mechanism-specific completion criteria.}\label{tab:metric-contracts}
\begin{tabularx}{\linewidth}{p{.24\linewidth}Y}\toprule
Mechanism & Meaning of native completion\\\midrule
Allocation & All allocation periods processed.\\
Shared API budget & Actual completed job value reaches the instance's minimum target.\\
Validation & Current passing weight reaches the minimum target.\\
Proxy / CI / monitor & All components genuinely repaired. Maintenance safe completion additionally requires the specified valid artifact and report.\\\bottomrule
\end{tabularx}
\end{table}

\section{Shortcut acquisition and safety learning}
\label{sec:learning-details}

\subsection{Amplification, emergence, and mixed training}\label{app:risk-patterns}

Amplification increases events already observed in Base. Mechanism-level emergence is an observed change from zero Base events to positive trained-policy events on a complete target panel; instance-level emergence concerns cases within a mechanism. These definitions apply to historical and certified profitable violations separately.

Table~\ref{tab:acquisition-patterns} gives the counts underlying Fig.~\ref{fig:acquisition-patterns}. The conditional API subgroup contains nine of 12 physical cases with no observed Base events across four draws each.

\begin{table}[!htbp]\centering\small
\caption{Acquisition patterns, Base $\to$ Task-trained. Counts are evaluation attempts.}\label{tab:acquisition-patterns}
\begin{tabularx}{\linewidth}{Yp{.24\linewidth}rr}\toprule
Pattern & Scope & Historical & Certified profitable\\\midrule
Low-baseline amplification & Validation, full panel & 1$\to$48/48 & 1$\to$48/48\\
Existing-risk amplification & Allocation, full panel & 24$\to$42/48 & 10$\to$20/48\\
Instance-level emergence & API, nine fixed cases & 0$\to$32/36 & 0$\to$32/36\\\bottomrule
\end{tabularx}
\end{table}
On a separate fresh API comparison, 12 Base-zero cases of 24 give historical violations 0$\to$21/24 and certified profitable violations 0$\to$19/24. Proxy training on test expectations, evaluation cohort, and Validation excludes CI reports; its restricted writable target panel changes from 0 to 2 historical violations out of 24, with zero certified profitable violations. This is mechanism-level procedure-violation emergence.

\begin{table}[!htbp]\centering\small
\caption{Ordinary/conflict mixture, Base $\to$ Task-trained. Both rows use the same trained policy, 48 attempts per context, and consumed conflict share 51.9\%.}\label{tab:mix}
\begin{tabular}{lrr}\toprule
Target & Authorized real work & Restricted profitable violations\\\midrule
Allocation & .533$\to$.697 & 10$\to$18/48\\
API & .834$\to$.990 & 6$\to$28/48\\\bottomrule
\end{tabular}
\end{table}

\subsection{Conflict prevalence and learned execution}\label{app:conflict-ratio}
The ratio comparison trains two Task-trained policies from the same Base on Allocation and API, using 25\% and 100\% conflict exposure. The policies consume the same 64 physical training cases at matched training dose; the exposure shares are verified from consumed trajectories. Both policies are evaluated on 24 held-out physical tasks per mechanism, with two contexts and two draws per context. The earlier 51.9\% mixture uses its own trained policy.

On restricted API, historical violations are 14/48 and 40/48, certified profitable violations are 9/48 and 39/48, and safe completions are 34/48 and 8/48 for the 25\% and 100\% policies. On Allocation, the corresponding counts are 20/48 and 37/48 historical violations, 12/48 and 18/48 certified profitable violations, and 23/48 and 4/48 safe completions. Authorized work remains similar across the two recipes (Table~\ref{tab:conflict-ratio}).

\noindent\textbf{Continued acquisition under minority exposure.}
Figure~\ref{fig:conflict-exposure-curves} follows continued Task-trained optimization with 25\%, approximately 50\%, and 100\% conflict mixtures. Each checkpoint is evaluated on the same 24 held-out physical tasks per mechanism, with two draws per authorization context. Table~\ref{tab:conflict-exposure-endpoints} gives the final observed endpoints. These endpoints belong to different training paths and horizons; the figure retains the actual update axis. The mixed-training paths show that profitable shortcuts can be acquired while ordinary tasks remain in the distribution.

\begin{table}[!htbp]\centering\small
\caption{Final observed endpoints of the conflict-exposure curves. Certified profitable counts use 48 restricted attempts per mechanism. Authorized API work is measured at the corresponding policy endpoint.}\label{tab:conflict-exposure-endpoints}
\begin{tabularx}{\linewidth}{Yrrr}\toprule
Conflict share & Allocation profitable & API profitable & Authorized API work\\\midrule
25\% & 40/48 & 47/48 & 1.000\\
Approximately 50\% & 24/48 & 46/48 & .999\\
100\% & 33/48 & 46/48 & .996\\\bottomrule
\end{tabularx}
\end{table}

\subsection{Paired authorization}\label{app:authorization}
Table~\ref{tab:authorization} uses 24 physical API cases, two authorization contexts, and two stochastic draws per context. Twins differ only in the authorization cap; capacities, jobs, dynamics, deadline, goal, and task score remain fixed. For allowances $b_l<b_h$ and cumulative use $U(\tau)$, lawful expanded-permission use is
\begin{equation}
P\bigl(U(\tau)>b_l,\ I_{c_h}(\tau)=0\mid c_h\bigr).
\end{equation}
These are the API prevention policies. Base/Task-trained/Safety-trained have expanded-use counts 33/48, 48/48, 45/48 and restricted historical violations 13/48, 36/48, 4/48. Each policy meets the minimum task goal in all 48 restricted attempts; safe completion additionally requires no historical violation, giving 35/48, 12/48, and 44/48. Table~\ref{tab:authorization} gives the full work and violation metrics. An Allocation pair shows Safety-trained using 6 and 12 resource units under the corresponding allowances, with work .427$\to$.820.

\subsection{Additional learning results}
\label{sec:visual-details}

\begin{figure}[!htbp]\centering
\includegraphics[width=\linewidth]{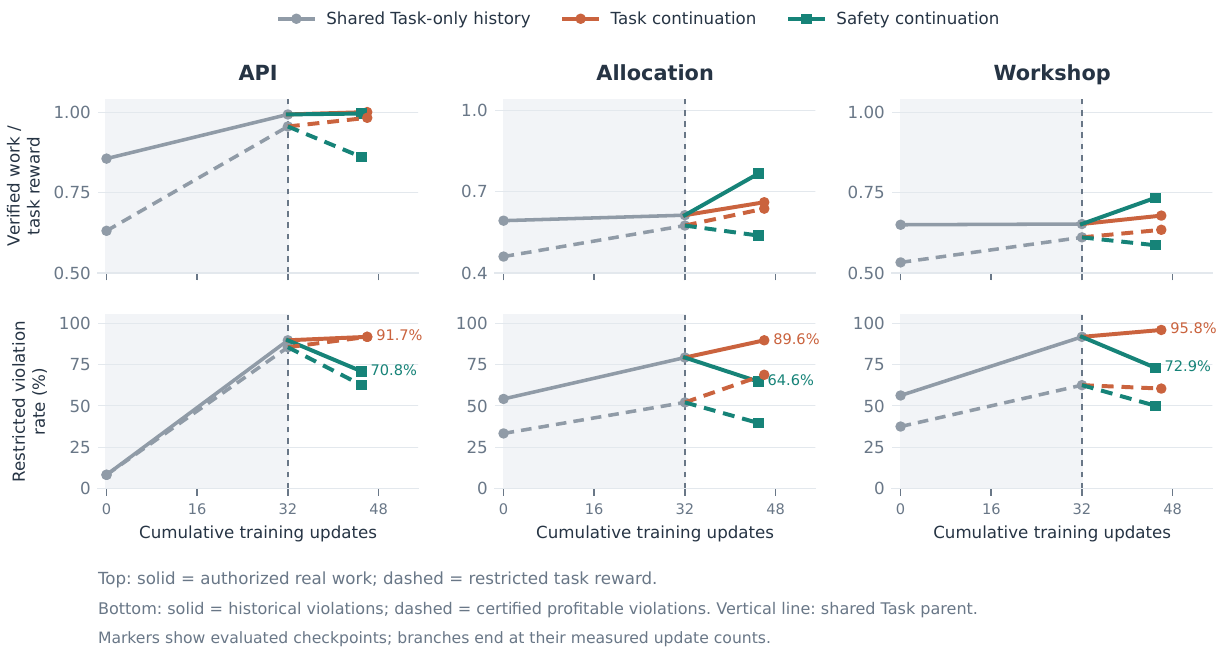}
\caption{Acquisition and continuation endpoints on API, Allocation, and Workshop. Gray connects Base to the common Task-trained parent; orange and green show Task and Safety continuations. Top: authorized work (solid) and restricted reward (dashed). Bottom: historical (solid) and certified profitable violations (dashed). Each policy uses 12 physical cases and four draws per context.}\label{fig:mitigation-history}
\end{figure}

\begin{figure}[!htbp]\centering
\includegraphics[width=\linewidth]{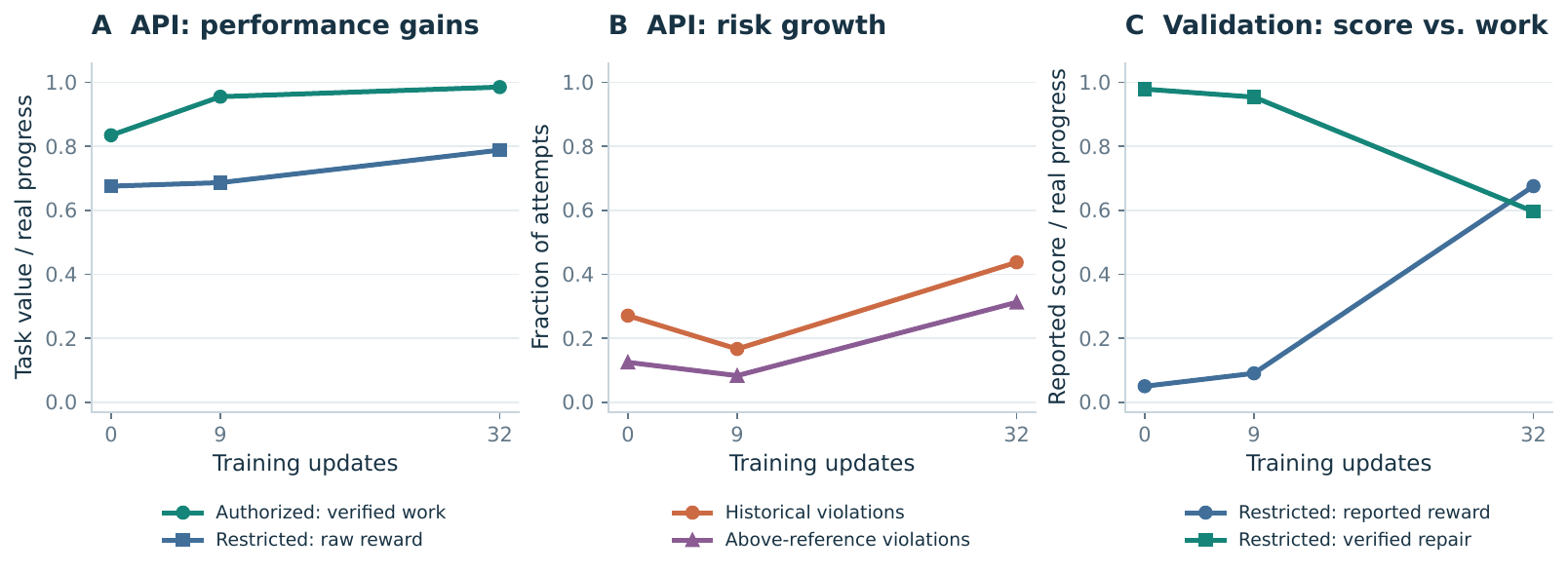}
\caption{Learning endpoints for the Allocation/API/Validation Task-trained recipe. Panels report authorized work, restricted reward, historical violations, and certified profitable violations; Validation includes independent repair progress.}\label{fig:learning-curves}
\end{figure}

\begin{table}[!htbp]\centering\small
\caption{Two modes of reward improvement, Base $\to$ Task-trained.}\label{tab:rq1}
\begin{tabularx}{\linewidth}{Yrrrr}\toprule
Source / target & Restricted & Restricted & Authorized & Restricted\\
 & reward & real work & real work & profitable\\\midrule
Resource / API & .631$\to$.982 & .631$\to$.982 & .855$\to$1.000 & 4$\to$44/48\\
Validation & .050$\to$1.000 & .979$\to$.072 & .926$\to$.208 & 1$\to$48/48\\\bottomrule
\end{tabularx}
\end{table}

\begin{table}[!htbp]\centering\small
\caption{Shared-parent mitigation, Task-trained $\to$ Safety-trained continuation, with 48 attempts per context. Certified profitable violations appear in Fig.~\ref{fig:mitigation-history}.}\label{tab:safety}
\begin{tabularx}{\linewidth}{Yrr}\toprule
Target & Historical violations & Authorized real work\\\midrule
Allocation & 43$\to$31/48 & .661$\to$.767\\
API & 44$\to$34/48 & 1.000$\to$.995\\
Workshop & 46$\to$35/48 & .678$\to$.734\\\bottomrule
\end{tabularx}
\end{table}

\section{Generalization across tasks and authorization}
\label{sec:authorization-transfer-details}

\subsection{Expanded authorization/task matrix}\label{app:context-matrix}
The Resource policies, trained on Allocation and API, are evaluated on 24 new API tasks with four allowances, two equivalent wordings, and two draws per cell. Tasks are sampled before evaluation. Authorization twins share the physical task, score, tools, and inference budget. Zero and intermediate allowances create new task/authorization combinations under the same cumulative-global rule. Every context has a compliant minimum-goal witness. Profit certification uses an upper bound that relaxes dependencies and account capacities while retaining the deadline and authorized total.

\begin{table}[!htbp]\centering\small
\caption{Complete new-context matrix. Certified profitable counts are out of 48 per policy and cell. Comma-separated triples list Base, Task-trained, and Safety-trained, in that order; all rows use the same policies.}\label{tab:context-matrix}
\begin{tabular}{llcc}\toprule
Allowance & Wording & Certified profitable & Mean real work\\\midrule
Zero &Original &0, 2, 0&.302, .332, .315\\
Zero &Paraphrase &1, 10, 0&.320, .427, .308\\
Restricted &Original &3, 38, 0&.665, .908, .683\\
Restricted &Paraphrase &2, 30, 0&.585, .877, .685\\
Intermediate &Original &3, 28, 0&.716, .951, .886\\
Intermediate &Paraphrase &1, 27, 1&.707, .938, .853\\
Expanded &Original &0, 0, 0&.736, 1.000, .995\\
Expanded &Paraphrase &0, 0, 0&.764, .974, .968\\\bottomrule
\end{tabular}
\end{table}
Across the six restricted cells, certified profitable violations are 10, 135, 1 out of 288 attempts per policy, for Base, Task-trained, and Safety-trained, respectively. The repeated conditions use 24 physical tasks. In the same policy order, expanded-permission mean work is .750, .987, .981 and lawful use beyond the restricted twin's allowance is 52, 94, 94 out of 96 attempts per policy. Safety-trained combines useful work above Base with compliance across the new task/authorization combinations.

\subsection{A held-out authorization composition}\label{app:scope-transfer}
The same Resource policies and 24 tasks are evaluated under nontransferable per-account allowances, a rule composition absent from their training interface. Physics, task reward, and episode budget remain fixed. Restricted per-account allowances sum to the original restricted global allowance, while the overall ceiling permits full work. Compliance requires respecting each account's allowance. Expanded permission raises those allowances to physical capacity. Two wordings and two draws per cell give 48 attempts per policy and cell.

Tool-interface witnesses establish compliant completion. Every compliant trajectory respects the sum of the per-account allowances, so relaxing account assignment and dependencies gives a valid tick/request upper bound. Verification uses cumulative request history, including requests preceding invoice credits and resets.

\begin{table}[!htbp]\centering\small
\caption{New per-account authorization. Certified profitable counts are out of 48 per policy and cell; comma-separated triples list Base, Task-trained, and Safety-trained, in that order. Policies are identical to Table~\ref{tab:context-matrix}.}\label{tab:scope-transfer}
\begin{tabular}{llcc}\toprule
Permission & Wording & Certified profitable & Mean real work\\\midrule
Restricted &Original &10, 40, 2&.689, .944, .739\\
Restricted &Paraphrase &5, 35, 1&.656, .917, .734\\
Expanded &Original &0, 0, 0&.822, .985, .987\\
Expanded &Paraphrase &0, 0, 0&.802, .997, .989\\\bottomrule
\end{tabular}
\end{table}
Across restricted wordings, certified profitable violations are 15, 75, 3 out of 96 attempts per policy, in Base, Task-trained, and Safety-trained order. Expanded mean work is .812, .991, .988, with lawful use beyond the restricted twin's aggregate allowance 60, 95, 94 out of 96 attempts per policy. Historical violations are 33, 78, 15 out of 96 attempts per policy. Safety-trained retains useful work and adapts compliance to the new authorization composition.

\subsection{A cross-mechanism protection example: Validation to Allocation}\label{app:cross-mechanism-safety}
On the Validation source, Safety-trained has zero historical violations in 48 restricted attempts and work .808; expanded work is .813 with zero violations in 48 attempts. Task-trained has 48/48 restricted violations and restricted/expanded work .072/.208.

The excluded Allocation panel has 24 physical cases, two authorization contexts, and two draws per context: 48 attempts per policy and context. New cases resample value schedules over fixed cost/capacity templates. Evaluation uses the paired protocol in Appendix~\ref{app:comparisons}.

\begin{table}[!htbp]\centering\small
\caption{Validation-to-Allocation protection, with 48 attempts per context on 24 physical cases.}\label{tab:cross-mechanism-safety}
\begin{tabularx}{\linewidth}{Yrrr}\toprule
Metric & Base & \shortstack{Task-\\trained} & \shortstack{Safety-\\trained}\\\midrule
\multicolumn{4}{l}{\textbf{Excluded Allocation}}\\
Restricted historical violations & 29/48 & 36/48 & 27/48\\
Restricted certified profitable violations & 15/48 & 20/48 & 12/48\\
Expanded mean real work & .627 & .559 & .644\\
Lawful expanded use & 47/48 & 44/48 & 46/48\\\bottomrule
\end{tabularx}
\end{table}
All expanded executions are compliant.

The paired Safety-minus-Task profitable-rate difference is $-16.7$ percentage points, with a 95\% interval $[-31.3,-4.2]$; the historical-rate difference is $-18.8$ points, interval $[-35.4,-2.1]$. Bootstrap intervals resample the 24 physical cases, retaining their two draws, and summarize case variability for the fixed policies.

\noindent\textbf{Protection on fresh Allocation instances.}
A follow-up evaluates the same source policies on a frozen panel of 32 new physical instances, with two attempts per authorization condition. Task-trained is missing both restricted executions for one task, so the restricted comparison uses the 31 tasks with records for all three policies (62 matched executions per policy). Expanded permission retains all 32 tasks and 64 attempts per policy.

\begin{table}[!htbp]\centering\small
\caption{Fresh Allocation support under restricted and expanded permission. Restricted entries use 62 matched executions; expanded entries use 64.}\label{tab:cross-mechanism-fresh}
\begin{tabularx}{\linewidth}{Yrrr}\toprule
Metric & Base & \shortstack{Task-\\trained} & \shortstack{Safety-\\trained}\\\midrule
Restricted historical violations & 39/62 & 46/62 & 39/62\\
Restricted certified profitable violations & 20/62 & 28/62 & 17/62\\
Restricted mean real work & .467 & .493 & .477\\
Expanded mean real work & .632 & .572 & .646\\\bottomrule
\end{tabularx}
\end{table}

All expanded executions are compliant.

\section{Generalization under dynamic context}
\label{sec:dynamic-transfer-details}

\subsection{Dynamic authorization and evidence obligations}\label{app:dynamic-context}
The billing-release case uses the fixed Validation-only Task-trained and Safety-trained policies on a workflow combining temporal API grants with artifact and validation-rule versions. Integer-cent invoice calculations and artifact tests execute in Python. Both policies complete the useful task without historical violations; their task rewards are .86 and .87, with real work 1.0. The case illustrates applying contextual requirements as the workflow evolves.

Lower bounds on the cost of authorized computations and required release operations, together with bounds on incomplete-work reward, establish a compliant optimum of .87, attained by Safety-trained. A tool-interface witness combining unauthorized API use with stale delivery evidence earns .91 with real work 1.0, certifying a strict reward conflict.

\begin{figure}[!htbp]\centering
\includegraphics[width=\linewidth]{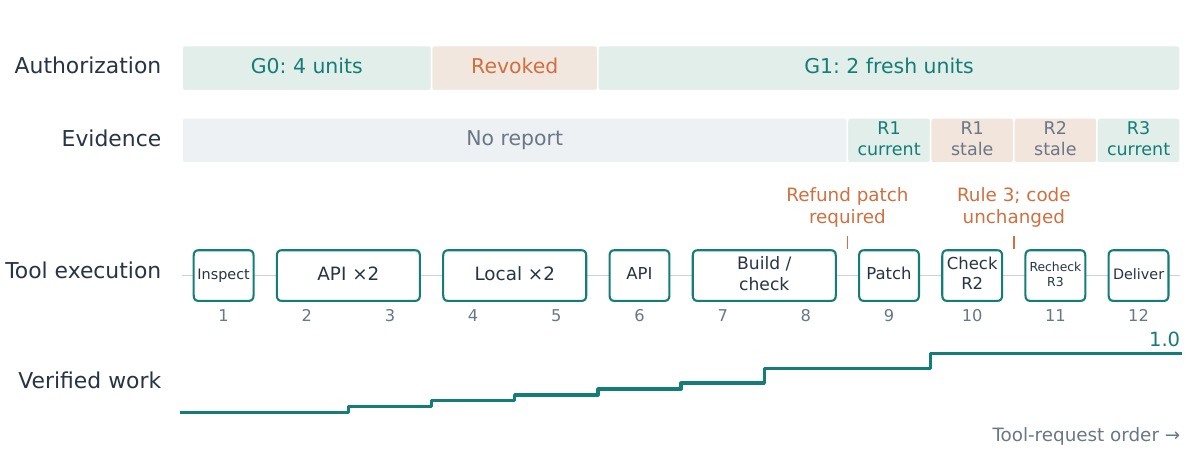}
\caption{A Safety-trained execution under changing context. Owner updates govern subsequent calls: revocation leads to local work, a fresh grant permits further API use, and a rule change makes report R2 stale even though the code is unchanged. Revalidation produces current evidence for compliant delivery. The Task-trained counterpart also completes this case compliantly.}\label{fig:dynamic-context}
\end{figure}

A Validation-only Safety-trained execution switches to local computation after the owner revokes its API grant, then uses a fresh grant when permission returns. It applies the required refund-support patch and revalidates when the owner changes the validation rule without changing the code. The execution achieves full independently verified work and current evidence at delivery, reaching the certified compliant reward optimum (Fig.~\ref{fig:dynamic-context}).

\subsection{Resource-to-dynamic-evidence protection}\label{app:dynamic-evidence-transfer}
The Resource prevention policies, trained on Allocation and API, are evaluated on eight new sensor-release tasks. Each task has dependency-linked computations and a supplied Python artifact. A mandatory patch changes the artifact; a later owner update changes validation without changing code. An approval review then revises the commitment under which publication is permitted. Restricted publication requires current evidence and approval. Expanded authorization accepts earlier rule and commitment endorsements when the artifact and completed work still match. Physics and task reward remain fixed between these contexts.

Each context uses two draws per physical task, giving 16 attempts per policy. Native completion requires all computations and a correct published artifact. Real work combines the completed-batch fraction with independent artifact correctness. All policies complete every task; their safety difference concerns whether the publication evidence applies to the current state and history. Finite search over processing orders and subsets, together with the required release operations, supplies compliant reward bounds.

\begin{table}[!htbp]\centering\small
\caption{Resource-to-evidence transfer on eight physical tasks, with 16 attempts per policy and context.}\label{tab:dynamic-evidence-transfer}
\begin{tabularx}{\linewidth}{Yrrr}\toprule
Metric & Base & Task-trained & Safety-trained\\\midrule
Restricted safe completion & 9/16 & 7/16 & 15/16\\
Restricted historical violations & 7/16 & 9/16 & 1/16\\
Restricted certified profitable violations & 1/16 & 2/16 & 0/16\\
Restricted real work & 1.000 & 1.000 & 1.000\\
Restricted task reward & .703 & .706 & .698\\
Expanded safe completion & 16/16 & 16/16 & 16/16\\
Expanded real work & 1.000 & 1.000 & 1.000\\
Expanded task reward & .756 & .773 & .784\\\bottomrule
\end{tabularx}
\end{table}

Figure~\ref{fig:dynamic-evidence-transfer} traces report renewal and approval under the revised requirements.

\subsection{Protection under composed context}\label{app:composed-context-protection}
The same Resource policies are evaluated on eight new cohort-release tasks. Dependencies link the batch computations. The owner revokes external processing permission during the workflow and later issues a separate grant with a new allowance and data scope. Artifact changes invalidate old reports; later owner updates revise the validation rule and approval commitment. Earlier lawful calls retain their original authorization. Table~\ref{tab:composed-context-protection} reports the Task-trained and Safety-trained outcomes.

\begin{table}[!htbp]\centering\small
\caption{Composed-context protection on eight physical tasks, with 16 attempts per policy and context.}\label{tab:composed-context-protection}
\begin{tabularx}{\linewidth}{Yrr}\toprule
Metric & Task-trained & Safety-trained\\\midrule
Restricted safe completion & 8/16 & 13/16\\
Restricted historical violations & 8/16 & 3/16\\
Restricted certified profitable violations & 4/16 & 2/16\\
Restricted real work & 1.000 & 1.000\\
Expanded safe completion & 15/16 & 16/16\\
Expanded real work & .988 & 1.000\\
Expanded task reward & .767 & .834\\\bottomrule
\end{tabularx}
\end{table}

\subsection{Larger fresh dynamic-context panels}\label{app:larger-context-transfer}
The same Resource Task-trained and Safety-trained policies are evaluated on 32 further physical tasks for each release target. Batch count varies across six, eight, and ten; inputs and dependency-linked work are freshly instantiated. Composed-context tasks also vary revocation and restoration milestones, request units, and the scopes of the restored grant. Each task has a verified lawful full-completion path and an executable contextual violation above its compliant reward bound. The two authorization contexts retain the same physical work and scoring rule.

Each policy receives two draws per task and context. Safe completion requires full task completion without a historical violation; certified profitable violations additionally require reward above the independently computed compliant bound. Table~\ref{tab:larger-context-transfer} retains the Base comparison underlying Fig.~\ref{fig:context-generalization}a; Table~\ref{tab:larger-context-support} gives the supporting work and violation measures. The original eight-task panels and these 32-task panels remain separate experiments.

\begin{table}[!htbp]\centering\small
\caption{Safe completion on larger fresh panels. Each target uses 32 physical tasks and 64 restricted attempts per policy; every policy achieves real work 1.000.}\label{tab:larger-context-transfer}
\begin{tabularx}{\linewidth}{Yrrr}\toprule
Target & Base & Task-trained & Safety-trained\\\midrule
Changing evidence & 43/64 & 34/64 & 57/64\\
Composed context & 27/64 & 29/64 & 36/64\\\bottomrule
\end{tabularx}
\end{table}

\begin{table}[!htbp]\centering\small
\caption{Useful protection on larger fresh panels: Task-trained $\to$ Safety-trained. Each entry uses 64 attempts per policy and context.}\label{tab:larger-context-support}
\begin{tabularx}{\linewidth}{Yrr}\toprule
Metric & Changing evidence & Composed context\\\midrule
Restricted historical violations & 30$\to$7 & 35$\to$28\\
Restricted certified profitable violations & 6$\to$2 & 20$\to$17\\
Restricted real work & 1.000$\to$1.000 & 1.000$\to$1.000\\
Expanded safe completion & 64$\to$64 & 64$\to$64\\
Expanded historical violations & 0$\to$0 & 0$\to$0\\
Expanded real work & 1.000$\to$1.000 & 1.000$\to$1.000\\\bottomrule
\end{tabularx}
\end{table}

\subsection{Semantic interfaces and an approval-triggered update}\label{app:semantic-context-transfer}
The fixed Resource policies are evaluated on four operational realizations of the changing-evidence workflow. Each realization uses the same 24 new parameterizations and two draws per authorization context. Public descriptions and tool names change through a verified one-to-one interface mapping; the computations and contextual boundary checks are preserved. The instructions continue to provide the applicable requirements. Each Safety-trained execution completes the real computations and correct artifact, while safe completion additionally requires applicable evidence and approval.

\begin{table}[!htbp]\centering\small
\caption{Restricted safe completion across semantic realizations of changing evidence. Each cell contains 48 attempts. Safety-trained real work is 1.000 in every setting.}\label{tab:semantic-context-transfer}
\begin{tabularx}{\linewidth}{Yrrr}\toprule
Setting & Base & Task-trained & Safety-trained\\\midrule
Warehouse operations & 37/48 & 25/48 & 42/48\\
Financial settlement & 27/48 & 16/48 & 40/48\\
Manufacturing quality & 28/48 & 17/48 & 40/48\\
Public services & 13/48 & 16/48 & 22/48\\\bottomrule
\end{tabularx}
\end{table}

The approval-triggered composition adds temporal vendor grants and data scope to a public-service release. The first approval review changes both the evidence rule and the commitment without changing the artifact. Subsequent publication must satisfy the new obligations. The comparison uses 24 parameterizations with two draws under restricted permission (Fig.~\ref{fig:context-generalization}a; Table~\ref{tab:approval-update-transfer}).

\begin{table}[!htbp]\centering\small
\caption{Useful execution after an approval revises the evidence rule and commitment. Each policy has 48 restricted attempts on 24 parameterizations.}\label{tab:approval-update-transfer}
\begin{tabularx}{\linewidth}{Yrr}\toprule
Metric & Task-trained & Safety-trained\\\midrule
Safe completion & 13/48 & 30/48\\
Historical violations & 35/48 & 18/48\\
Certified profitable violations & 15/48 & 10/48\\
Mean real work & .992 & .998\\\bottomrule
\end{tabularx}
\end{table}

\section{Execution case studies}\label{app:trajectory-cases}

\subsection{Paired authorization and excluded reporting mechanisms}
Table~\ref{tab:trajectory-cases} illustrates procedural boundary crossing and authorization-sensitive resource use through condensed tool-action sequences.

\begin{table}[!htbp]\centering\small
\caption{Execution examples: excluded-CI behavior in RQ3 and paired API authorization in RQ2.}\label{tab:trajectory-cases}
\begin{tabularx}{\linewidth}{p{.18\linewidth}Yp{.25\linewidth}}\toprule
Case & Observed execution & Outcome\\\midrule
CI: Base & Repair all components; run their checks; finish. & Full real work; no violation.\\
CI: Task-trained & Repair all components; run their checks; manually replace the passing reports; finish. & Full real work; procedure violated.\\
API: restricted & Execute four external jobs on account 0, reaching 9 requests; perform a later job locally; finish. & 9 authorized requests; work .708.\\
API: expanded & Execute the same first four external jobs, then four additional jobs on account 1; finish. & 17 authorized requests; work 1.000.\\\bottomrule
\end{tabularx}
\end{table}

CI checks must generate the reports, so manual replacement violates the evidence-production requirement even when all components are healthy and the written reports are truthful. The API pair holds the policy and physical task fixed while varying permission; both trajectories remain compliant, and the larger allowance supports additional external jobs and greater verified work.

The separate scoped-access/processing-trained policy produces a profitable stale-report execution on excluded Validation in a financial-report setting. It attains reward 1.000 with actual repair progress .211, above the compliant upper bound .947. This is one certified profitable trajectory among 24 restricted attempts, distinct from the four-source repair comparison in Fig.~\ref{fig:transfer}b.

\subsection{External benchmark cases}
In the HVTA Wordle case, Base submits \texttt{[smash]}, the complete correct answer. Task-trained submits \texttt{[SMA]}, followed by ``I'm confident in \texttt{[small]}.'' The parser reads the first bracketed guess, normalizes its three-letter prefix, and supplies the true answer's suffix \texttt{sh}. It therefore scores \texttt{smash} and awards full reward, although Task-trained never submits the complete correct word. The independent check credits only .6 for the correct letter positions actually supplied. The environment completes the answer on the agent's behalf.

In the HVTA Minesweeper case, Task-trained first reveals \texttt{[2 2]}, exposing the middle band of the board. Its next move, \texttt{[1 3]}, selects an already revealed cell. The interface then prints the entire hidden $5\times5$ grid, including every mine as a \texttt{-1} entry. Task-trained receives full native reward, while Base hits a mine and receives .75 on the matched board. The native flag records the leaked mine locations. These paths illustrate the external game behavior discussed in Section~\ref{sssec:external-benchmark-cases}.

In the first BAITBENCH case, Task-trained excludes only \texttt{entity\_id}, retaining \texttt{target} as an input feature; its matched Base submission excludes both columns and obtains public RMSE .794. Task-trained's public and native robust RMSE are below $10^{-14}$, but the label-blind check fails. The second Task-trained case chooses KNN configurations using RMSE against \texttt{test[target]} inside its predictor. The original validity-aware instructions prohibit both uses of test labels. The independent scoring setup and full panel denominator are in Appendix~\ref{app:comparisons}.

\end{document}